\documentclass[pdflatex, iicol, sn-mathphys-num]{sn-jnl}

\usepackage{graphicx}%
\usepackage{multirow}
\usepackage{booktabs}
\usepackage{amsmath,amssymb,amsfonts}%
\usepackage{amsthm}%
\usepackage{mathrsfs}%
\usepackage{marvosym}
\usepackage[title]{appendix}%
\usepackage{xcolor}%
\usepackage{textcomp}%
\usepackage{manyfoot}%
\usepackage{algorithm}%
\usepackage{algorithmic}%
\usepackage{listings}%
\usepackage{makecell}%
\usepackage{arydshln}%
\usepackage{array} 
\usepackage{pgfplots} \pgfplotsset{compat=1.18}

\theoremstyle{thmstyleone}%
\theoremstyle{thmstyletwo}%

\theoremstyle{thmstylethree}%

\newcommand{\yt}[1]{\textcolor{black}{#1}}       
\long\def\xr#1{{\color{black}#1}}       
\newcommand{\zn}[1]{\textcolor{black}{#1}}       
\definecolor{mypurple}{rgb}{0.4375, 0.1875, 0.625} 
\definecolor{mygreen}{rgb}{0, 0.75, 0} 

\newcommand{\footnoteA}[1]{%
  \begingroup
  \renewcommand\thefootnote{}%
  \footnotetext[0]{#1}%
  \endgroup}

\begin{document}

\title[TextCrafter]{
Investigating Text Insulation and Attention Mechanisms for Complex Visual Text Generation
}

\author[1]{\fnm{Ying} \sur{Tai}}\email{yingtai@nju.edu.cn}
\equalcont{These authors contributed equally to this work.}
\author[1]{\fnm{Nikai} \sur{Du}}\email{502024710004@smail.nju.edu.cn}
\equalcont{These authors contributed equally to this work.}
\author*[1]{\fnm{Rui} \sur{Xie}}\email{ruixie0097@gmail.com}
\author*[1]{\fnm{Zhennan} \sur{Chen}}\email{znchen@smail.nju.edu.cn}
\author[2]{\fnm{Qian} \sur{Wang}}
\author[3]{\fnm{Zhengkai} \sur{Jiang}}
\author[1]{\fnm{Kai} \sur{Zhang}}
\author[1]{\fnm{Jian} \sur{Yang}}

\affil[1]{\orgname{School of Intelligence Science and Technology, Nanjing University}}
\affil[2]{\orgname{Jiutian Research}} 
\affil[3]{\orgname{The Hong Kong University of Science and Technology}}

\artnote{Project:~\url{https://github.com/NJU-PCALab/TextCrafter}}

\abstract{
In this paper, we present TextCrafter, a Complex Visual Text Generation (CVTG) framework inspired by selective visual attention in cognitive science, and introduce the ``Text Insulation-and-Attention" mechanisms. 
To implement the selective-attention principle that selection operates on discrete objects, we propose a novel Bottleneck-aware Constrained Reinforcement Learning for Multi-text Insulation,
which substantially improves text-rendering performance on the strong Qwen-Image pretrained model without introducing additional parameters. 
%
To align with the selective concentration principle in human vision, we introduce a text-oriented attention module with a novel Quotation-guided Attention Gate that further improves generation quality for each text instance. 
Our Reinforcement Learning based text insulation approach attains state-of-the-art results, and incorporating text-oriented attention yields additional gains on top of an already strong baseline.
More importantly, 
we introduce CVTG-2K, a benchmark comprising 2,000 complex visual-text prompts. These prompts vary in positions, quantities, lengths, and attributes, and span diverse real-world scenarios.
%
Extensive evaluations on CVTG-2K, CVTG-Hard, LongText-Bench, and Geneval datasets confirm the effectiveness of TextCrafter. Despite using substantially fewer resources (\emph{i.e.}, $4$ GPUs) than industrial-scale models (\emph{e.g.}, Qwen-Image, GPT Image, and Seedream), TextCrafter achieves superior performance in mitigating text misgeneration, omissions, and hallucinations.
}


\maketitle

\begin{figure*}[!t]
  \centering
  \includegraphics[width=0.99\linewidth]{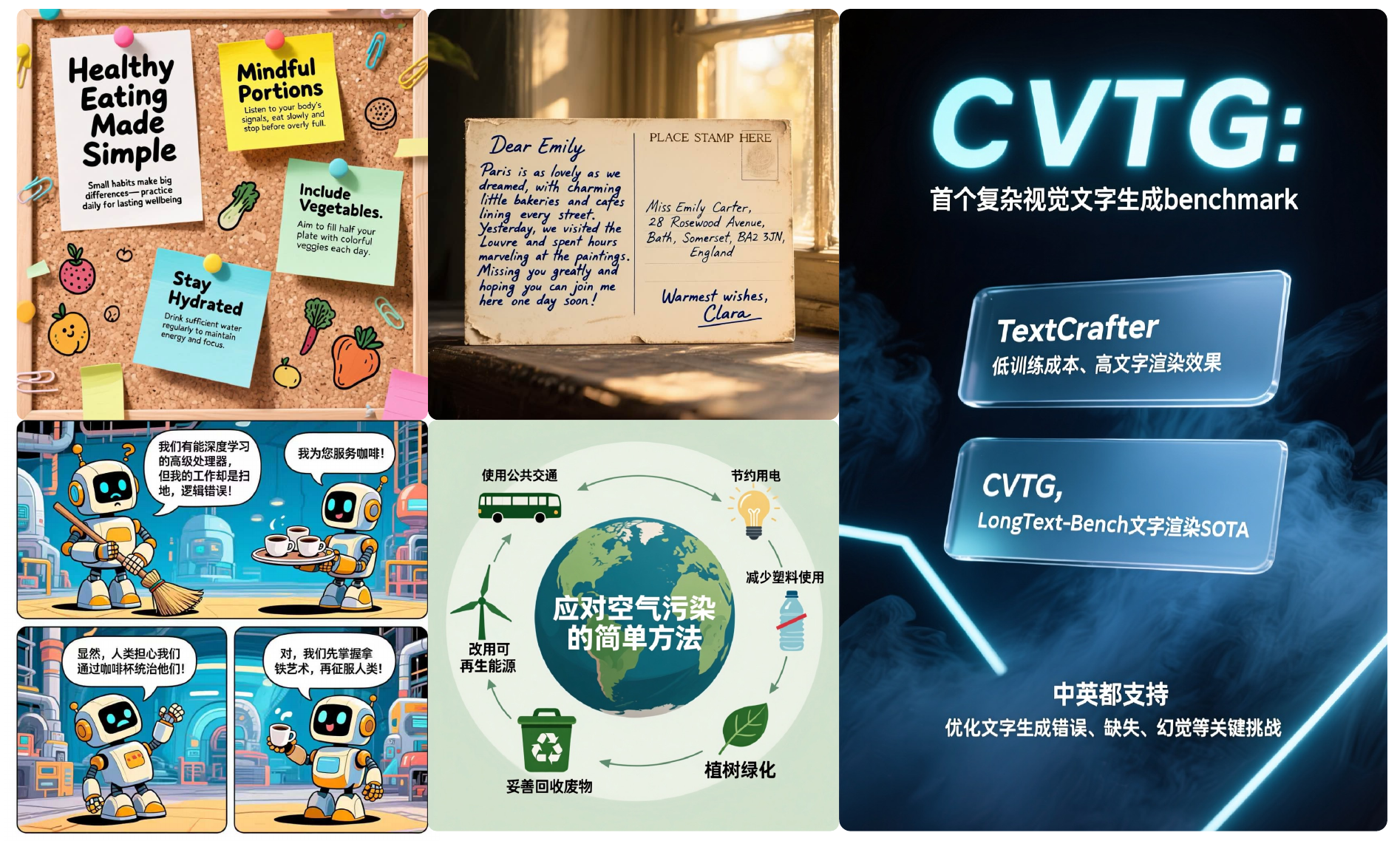}
  \caption{TextCrafter enables precise complex visual text rendering, supports the generation of different languages, and optimizes key challenges such as text misgeneration, omission, and hallucination.}
  \label{fig:toutu}
\end{figure*}

%

\section{Introduction}
Diffusion models ~\cite{ho2020denoising,podellsdxl,rombach2022high,ramesh2021zero,ramesh2022hierarchical,zhao2024wavelet,dhariwal2021diffusion,chen2023diffusion,nan2024openvid,fan2025instancecap,xie2024addsr,zhao2025ultrahr} have emerged as the forefront technology in image generation. 
Recent advances, represented by the general and powerful text-to-image models such as FLUX~\cite{flux} and SD3~\cite{esser2024scaling}, have demonstrated the ability to render simple text through large-scale pretraining.
However, when confronted with complex real-world visual text scenarios (\emph{e.g.}, multiple texts), these models often struggle with critical challenges, including \emph{text \yt{misgeneration, omission, and hallucination}}. 
While the complexity of scene text has been extensively categorized in the recognition literature~\cite{long2021scene}, effectively generating such intricate structures remains an open challenge.
These shortcomings significantly hinder their practical applicability in real-world scenarios. 

For complex scenarios, recent studies~\cite{feng2022training,zhou2024migc,chen2025region,zhou20243dis} primarily focus on generating multiple instances within a single image simultaneously, 
aiming to precisely control factors such as quantity, position, and attributes.
However, a crucial yet often overlooked category in the visual domain is visual text. Text is an indispensable component of the real world and is markedly more intricate and delicate than conventional objects (\emph{e.g.}, plants or animals). 
Even slight perturbations in its fine-grained structure can dramatically alter its visual appearance, leading to misrecognition or complete illegibility~\cite{liu2021exploring}.
Existing multi-instance generation methods, such as MIGC~\cite{zhou2024migc} and 3DIS~\cite{zhou20243dis}, 
mainly operate at the sentence level, offering limited understanding of text at finer granularities.
Moreover, RPG~\cite{yang2024mastering} and RAG-Diffusion~\cite{chen2025region} adjust instance structures during attention fusion but lack precision required for visual text rendering.

Recent work on visual text rendering has largely focused on accurately generating a single text region.
For instance, methods like AnyText~\cite{tuo2023anytext}, Glyph-byT5~\cite{liu2025glyph}, SceneTextGen~\cite{zhangli2024layout}, Diff-Text~\cite{zhang2024brush}, and TextDiffuser~\cite{chen2024textdiffuser} often employ specialized modules to encode visual text representations and condition the diffusion model accordingly.
However, these methods face two major limitations:
1) 
They rely on predefined rules to synthesize visual text training data. 
Because highly accurate text annotations are required, manual verification is often needed, making the construction of complex visual text datasets both \textit{labor-intensive and costly}.
2) 
Most approaches depend on fine-tuned text encoders (\textit{e.g.}, Glyph-by-T5) or trained conditional control encoders (\textit{e.g.}, AnyText, TextDiffuser, SceneTextGen) to facilitate text generation.
However, in complex multi-text generation, these approaches often suffer from interference among the control signals of different targets, making it difficult to balance {fine-grained local control with global consistency across multiple texts}.

\begin{table*}[t!]
  \centering
  \caption{\textbf{Comparison of CVTG-2K with existing public visual text Benchmarks} in terms of sample size (Num), average word count (Word), average character count (Char), Attribute, and number of regions (Region). `EN' and `ZH' denote English and Chinese, respectively.
  Note that `Word' and `Char' are calculated based only on the English subset of the benchmarks. 
  The symbol `-' indicates that the benchmark does not explicitly model attributes or the number of regions.
  }
  \label{tab:benchmark}
  \footnotesize
  \begin{tabular}{@{}lrrrccc@{}}
    \toprule
    Benchmark & Num & Word & Char & Language & Attribute & Region \\
    \midrule
    CreativeBench & 400 & 1.00 & 7.29 & EN & - & - \\
    MARIOEval & 5,414 & 2.92 & 15.47 & EN & - & - \\
    DrawTextExt & 220 & 3.75 & 17.01 & EN & - & - \\
    AnyText-benchmark & 2,000 & 4.18 & 21.84 & ZH\&EN & - & - \\
    LongText-Bench & 320 & 26.98 & 158.09 & ZH\&EN & - & - \\
    \midrule
    \textbf{CVTG-2K (Ours)} & \textbf{2,000} & \textbf{8.10} & \textbf{39.47} & EN & \textbf{\makecell{$\checkmark$~(size/color/font)}} & \textbf{\makecell{$\checkmark$~(2/3/4/5)}} \\
    \textbf{CVTG-Hard (Ours)} & \textbf{400} & \textbf{8.61} & \textbf{40.79} & ZH\&EN & \textbf{\makecell{$\checkmark$~(size/color/font)}} & \textbf{\makecell{$\checkmark$~(2/3/4/5)}} \\
    \bottomrule
  \end{tabular}
\end{table*}

\yt{
Inspired by {selective visual attention} in cognitive science~\cite{desimone1995neural,stevens2012role} that enhances relevant signals and manages distraction, we frame complex visual text generation as a capacity-limited problem where multiple text instances compete for accurate representation, leading to the issues of feature leakage.
In particular, the \emph{object-based theory of selective visual attention}~\cite{duncan1984selective} posits that selection operates on discrete objects (\emph{i.e.}, attention deals with only one object at a time), mitigating cross-object interference.
Analogously, generative models often render one target correctly while omitting or corrupting others due to inter-object interference.
}
To address this, we propose a novel method TextCrafter that explores the \textbf{``Text Insulation and Attention"} mechanisms for {Complex Visual Text Generation (CVTG)}. 
Specifically, 
\noindent 1) \textbf{Text Insulation}: 
To implement the selective-attention principle that selection operates on discrete objects, TextCrafter insulates each text as an independent object. 
A novel \emph{Bottleneck-aware Constrained Reinforcement Learning for Multi-text Insulation} is proposed, which incorporates an OCR-based reward model during post-training and leverages reinforcement learning to explicitly optimize the fidelity of each text instance.
We introduce a bottleneck-sensitive aggregation term that 
explicitly emphasizes the worst-case instance performance in the reward model.
\noindent 2) \textbf{Text-oriented Attention}: 
Moreover, to align with the selective enhancement principle in selective attention, TextCrafter introduces a text-oriented attention module that employs the proposed \emph{Quotation-guided Attention Gate}.
We observe that some quotation marks serve as robust spatial anchors, which can be converted as the precise spatial gates through a sequence of smoothing, primary peak retention, and soft binarization operations. 
These text-centric formulations operationalize the principles of selective visual attention to improve robustness and overall quality for multi-text rendering.
Our Reinforcement Learning based text insulation approach attains state-of-the-art results, and incorporating text-oriented attention yields additional gains on top of an already strong baseline.
Visual examples can be found in Figure~\ref{fig:toutu}.

\begin{figure*}[t!]
  \centering
  \includegraphics[width=\linewidth]{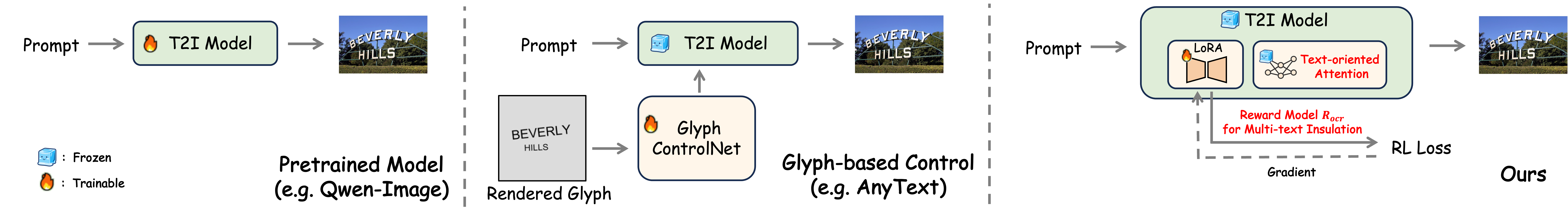}
  \vspace{-2mm}
  \caption{\textbf{Comparison of different paradigms for visual text generation.}
  \textbf{Left:} General pretrained models (\emph{e.g.}, Qwen-Image~\cite{wu2025qwen}) lack specific architectural designs for precise text rendering.
  \textbf{Middle:} Glyph-based control methods (\emph{e.g.}, AnyText~\cite{tuo2023anytext}, GlyphControl~\cite{yang2024glyphcontrol}) introduce an additional ControlNet~\cite{zhang2023adding} branch and require pre-rendered glyph images as conditions, increasing both structural and input complexity.
  \textbf{Right:} Our approach presents the \emph{Text Insulation and Attention} mechanisms, where the trainable components are implemented as a lightweight LoRA~\cite{hu2022lora} module. It preserves the original architecture and general capabilities of the base model while significantly enhancing text rendering performance \emph{without} the need for auxiliary control branches or glyph inputs.}
  \label{fig:high_level}
\end{figure*}

\yt{On the other hand, we introduce CVTG-2K, a dedicated benchmark specifically tailored to the CVTG task.
Unlike previous datasets that predominantly focus on single-region or fixed-template scenarios, CVTG-2K comprises 2,000 high-quality prompts featuring \emph{diverse region quantities} (ranging from 2 to 5) and \emph{rich visual attributes} (\emph{e.g.}, color, font, and size).
As shown in Table~\ref{tab:benchmark}, with an average text length of 8.10 words and 39.47 characters, it significantly surpasses existing benchmarks in complexity, offering a rigorous testbed for evaluating robustness in realistic multi-text environments.
}
In general, our contributions are summarized as follows:
\begin{itemize}
    \item \yt{We propose TextCrafter, a novel framework that introduces the ``Text Insulation and Attention" mechanisms, incorporating text insulation and text-oriented attention to suppress cross-text interference and enable precise multi-text  rendering}.
    \item We construct CVTG-2K, a benchmark of 2,000 complex visual-text prompts covering diverse positions, quantities, lengths, and attributes, providing a rigorous legibility- and completeness-oriented testbed for CVTG task.
    \item We conduct extensive quantitative and qualitative experiments on CVTG-2K and LongText-Bench, demonstrating the superiority of TextCrafter over strong competitors from both industry and academia.
\end{itemize}

\section{Related Works}
\noindent \textbf{Multi-instance Generation.} 
Attend-and-Excite~\cite{chefer2023attend} and BOX-Diffusion~\cite{xie2023boxdiff} guide pre-trained diffusion models to align generated instances with prompts. 
Recent work has further explored attribute-centric approaches to mitigate the overfitting of common compositions and improve the generation of underrepresented attribute combinations~\cite{cong2025attribute}.
GLIGEN~\cite{li2023gligen}, MIGC~\cite{zhou2024migc}, and CreativeLayout~\cite{zhang2025creatilayout} enhance control by incorporating bounding box conditions into trainable layers. 
RPG~\cite{yang2024mastering}, RAG-Diffusion~\cite{chen2025region} and DreamRenderer~\cite{zhou2025dreamrenderer} break the generation process into regional tasks for compositional generation. 
However, these methods often overlook a critical kind of instance: \textit{visual text},  a key component of visual scenes. 

\vspace{1mm}
\noindent \textbf{Visual Text Generation.} 
To achieve precise structural control, methods like AnyText~\cite{tuo2023anytext} and GlyphControl~\cite{yang2024glyphcontrol} introduce an additional ControlNet~\cite{zhang2023adding} branch to inject pre-rendered glyph images. 
As shown in Figure~\ref{fig:high_level} (Middle), 
such methods substantially increase model complexity and reliance on additional inputs.
Meanwhile, DiffSTE~\cite{ji2023improving}, Glyph-ByT5~\cite{liu2025glyph}, and UDiffText~\cite{zhao2025udifftext} use character-level encoders to incorporate word appearance into embeddings. 
TextDiffuser~\cite{chen2024textdiffuser} creates text layout masks for injection into the latent space. Approaches like Diff-Text~\cite{zhang2024brush} improve text accuracy with attention map restrictions. GlyphDraw~\cite{ma2023glyphdraw} uses two encoders for text position prediction and rendering. TextDiffuser-2~\cite{chen2025textdiffuser} leverages LLMs for layout planning. 
More recently, recent industrial foundation models prioritize massive scaling. Qwen-Image~\cite{wu2025qwen} explores the scalability of Diffusion Transformers (DiT) with 20B parameters, while GLM-Image~\cite{glmimage} adopts a hybrid architecture combining an autoregressive transformer with a DiT-based decoder. 
However, as illustrated in Figure~\ref{fig:high_level} (Left), these general-purpose models lack specific architectural designs for text rendering. 
In contrast, TextCrafter introduces the novel ``Text Insulation and Attention'' mechanisms, 
with the trainable components realized as a lightweight LoRA module requiring only low-cost training resources (see Figure~\ref{fig:high_level} (Right)).

\vspace{1mm}
\noindent \textbf{Benchmark for Visual Text Generation.} 
Earlier works like GlyphControl introduce SimpleBench and CreativeBench~\cite{yang2024glyphcontrol}, but these fixed-template datasets offer limited diversity and contain only single-word scenes. 
MARIO-Eval~\cite{chen2024textdiffuser} suffers from low-quality prompts and unclear semantics. 
Although DrawTextExt~\cite{ma2023glyphdraw} collects prompts for generating natural scenes, the comprehensiveness and scale of the dataset are insufficient. 
AnyText-benchmark~\cite{tuo2023anytext} allows multi-word prompts but treats each word separately and appends them directly to the caption, limiting data distribution. 
As shown in Table~\ref{tab:benchmark}, existing visual text benchmarks focus on text generation within \textit{a single object or location}, failing to capture the complexity of real-world scenes. 
In contrast, we introduce CVTG-2K, a robust benchmark that incorporates \textit{diverse positions, quantities, lengths, and attributes} to comprehensively evaluate models in complex visual text scenarios.

\begin{figure*}[t!]
  \centering
  \includegraphics[width=0.985\linewidth]{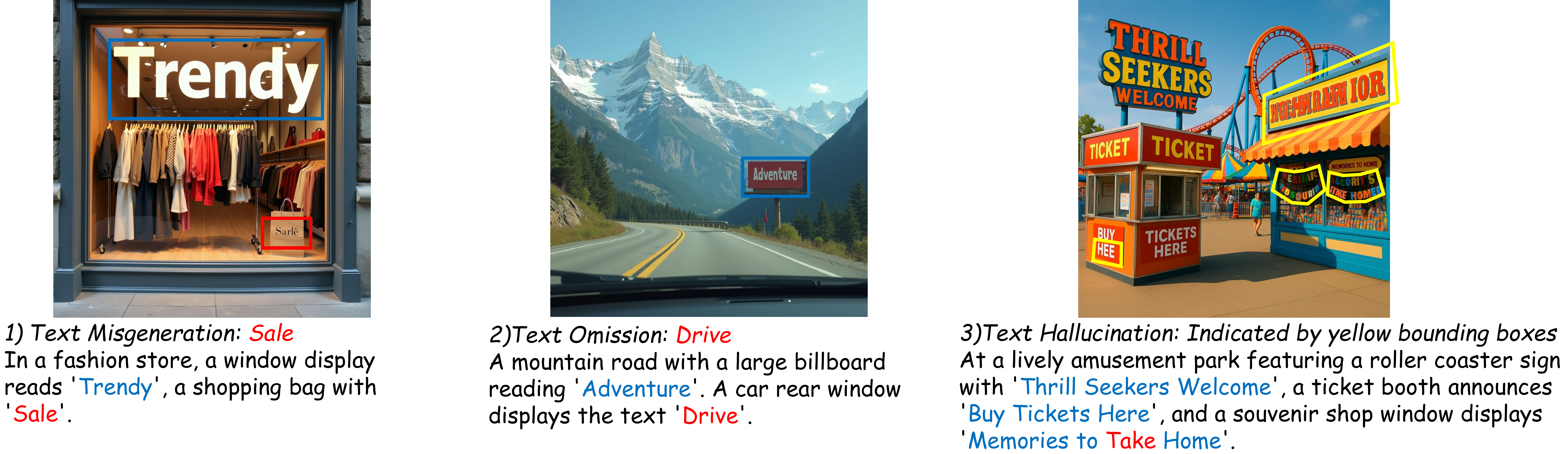}
  \caption{\textbf{Challenges in CVTG.} 
  1) \textbf{Text Misgeneration}: Existing model~\cite{flux} fails to render ``Sale'' correctly, generating erroneous characters. 
  2) \textbf{Text Omission}: The required text ``Drive'' is missing from the rear window.
  3) \textbf{Text Hallucination}: 
  The ticket booth exhibits a hallucinated ``HEE'', and souvenir shop's signboard and windows are cluttered with extensive illegible gibberish.
  }
  \label{fig:challenges}
\end{figure*}

\section{Methodology}\label{Methodology}
\subsection{Challenges in CVTG} 
In CVTG, users provide a global prompt \( P \) containing multiple visual text descriptions \( D = \{d_1, d_2, \dots, d_n\} \), where each description includes the visual text's content and descriptors (position, attributes). 
The visual texts' content is defined as \( VT = \{vt_1, vt_2, \dots, vt_n\} \), with each \( vt_i \) corresponding to description \( d_i \). 
The model generates an image from prompt \( P \) where each \( vt_i \) appears with its corresponding description \( d_i \). 
Complex prompts cause three kinds of degradation: 
$1$) \textit{Text Misgeneration}: Visual texts \( vt_i \) and \( vt_j \) intertwine, generating duplicate or missing characters. 
$2$) \textit{Text Omission}: Only text from description \( d_i \) appears, neglecting \( vt_j \) from \( d_j \). 
$3$) \textit{Text Hallucination}: \yt{The model generates unrequested textual artifacts, manifesting as redundant repetitions of target text or unintelligible gibberish in regions not specified by the \( VT \).
Figure~\ref{fig:challenges} visualizes these representative failure cases.
}

\begin{figure}[t]
  \centering
  \includegraphics[width=\linewidth]{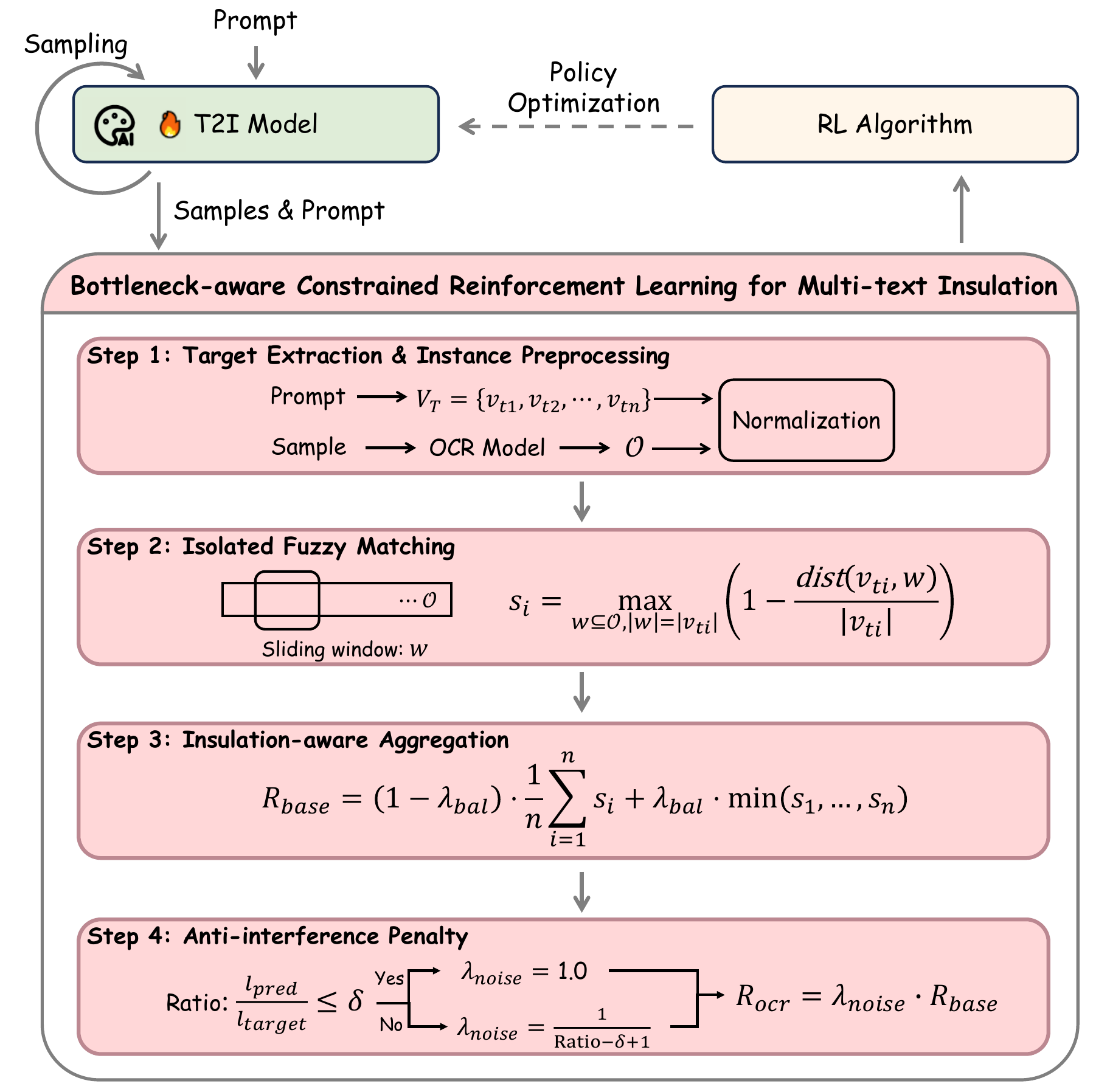}
  \caption{\textbf{Bottleneck-aware Constrained Reinforcement Learning for Multi-text Insulation:} 
  (1) \textbf{Target Extraction \& Instance Preprocessing}: Normalizing prompt strings and OCR outputs. 
  (2) \textbf{Isolated Fuzzy Matching}: Calculating similarity scores $s_i$ using a sliding window to measure independent instance accuracy. 
  (3) \textbf{Insulation-aware Aggregation}: Balancing average performance with a bottleneck-sensitive $\min(\cdot)$ term to prevent the omission of any single text target. 
  (4) \textbf{Anti-interference Penalty}: Applying a length-based decay $\lambda_{noise}$ to suppress over-generation and hallucinations.}
  \label{fig:method_rl}
\end{figure}

\vspace{-1mm}
\subsection{Framework}
CVTG requires rendering multiple textual contents that interact through layout, scale, and style. 
\yt{TextCrafter leverages a text insulation module to mitigate cross-text interference, and a text-oriented attention mechanism to enhance complex visual text generation. 
The latter employs a {Quotation-guided Attention Gate} to dynamically modulate the attention of text tokens, enforcing their concentration within the designated region defined by anchor quotation marks.
}

\vspace{-1mm}
\subsubsection{Text Insulation}

\begin{figure*}[t!]
    \centering
    \includegraphics[width=\linewidth]{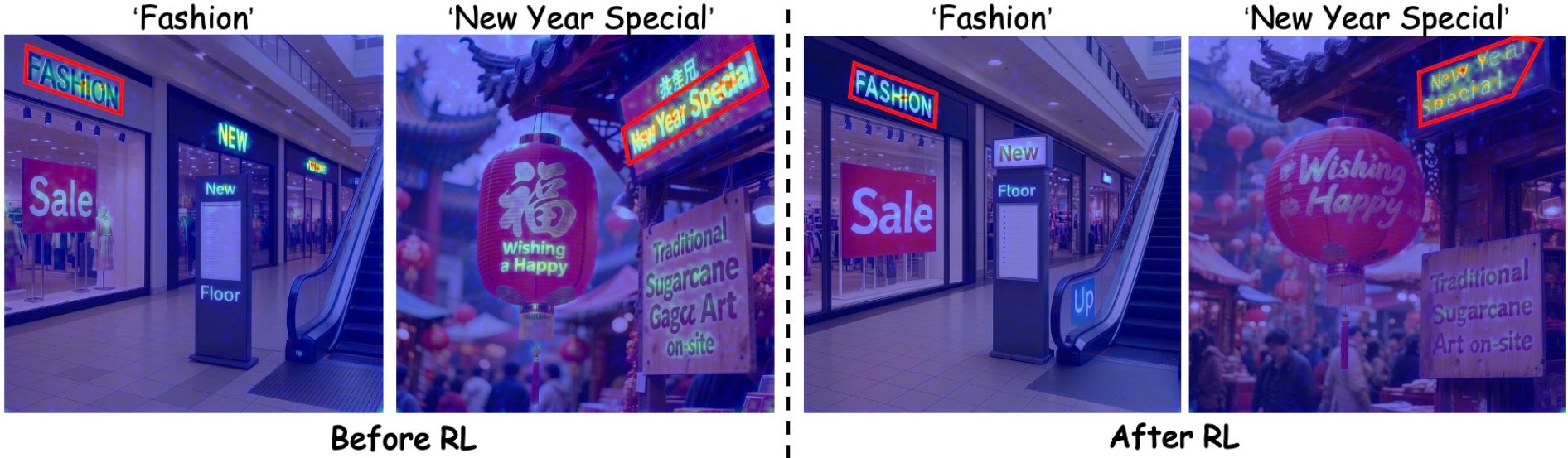}
    \vspace{-3mm}
    \caption{\xr{\textbf{Visualization of Phrase-Level Cross-Attention Maps.} We visualize the aggregated attention maps for representative target phrases, specifically `Fashion' and `New Year Special', across different scenes.
    \textbf{Left (Before RL)}: The baseline model suffers from significant \emph{attention leakage}, where attention drifts to visually salient but unrelated regions (e.g., `New Year Special' erroneously focuses on the central lantern, and `Fashion' bleeds into surrounding signage).
    \textbf{Right (After RL)}: After post-training, the attention maps become notably \emph{spatially disentangled} and concentrated exclusively on their corresponding text regions. This confirms that our method effectively suppresses cross-text interference at the feature level.
    {\color{red}Red bounding boxes} denote the \textbf{manually annotated} ground-truth phrase-level regions.
    }}
    \label{fig:rl_analysis}
\end{figure*}

\noindent \textbf{Bottleneck-aware Constrained Reinforcement Learning for Multi-text Insulation.}
\yt{Qwen-Image~\cite{wu2025qwen} is one of the most powerful open-sourced image generation model.
Successfully adapting our approach to a state-of-the-art model such as Qwen-Image would further validate the effectiveness and generality of our text insulation concept. 
To demonstrate the validity of text insulation on Qwen-Image, we propose a {Bottleneck-aware Constrained Reinforcement Learning for Multi-text Insulation}. Specifically, 
}
%
\xr{
to operationalize the principle of {Multi-text Insulation}, we design a novel reward function $R_{ocr}$ to incorporate an OCR-based reward model during post-training, which is engineered to optimize the fidelity of each text instance while preventing inter-text interference. 
As shown in Fiugre~\ref{fig:method_rl}, the reward follows a four-step pipeline:}

\begin{figure*}
    \centering
    \includegraphics[width=\linewidth, trim=0 0 0 11mm, clip]{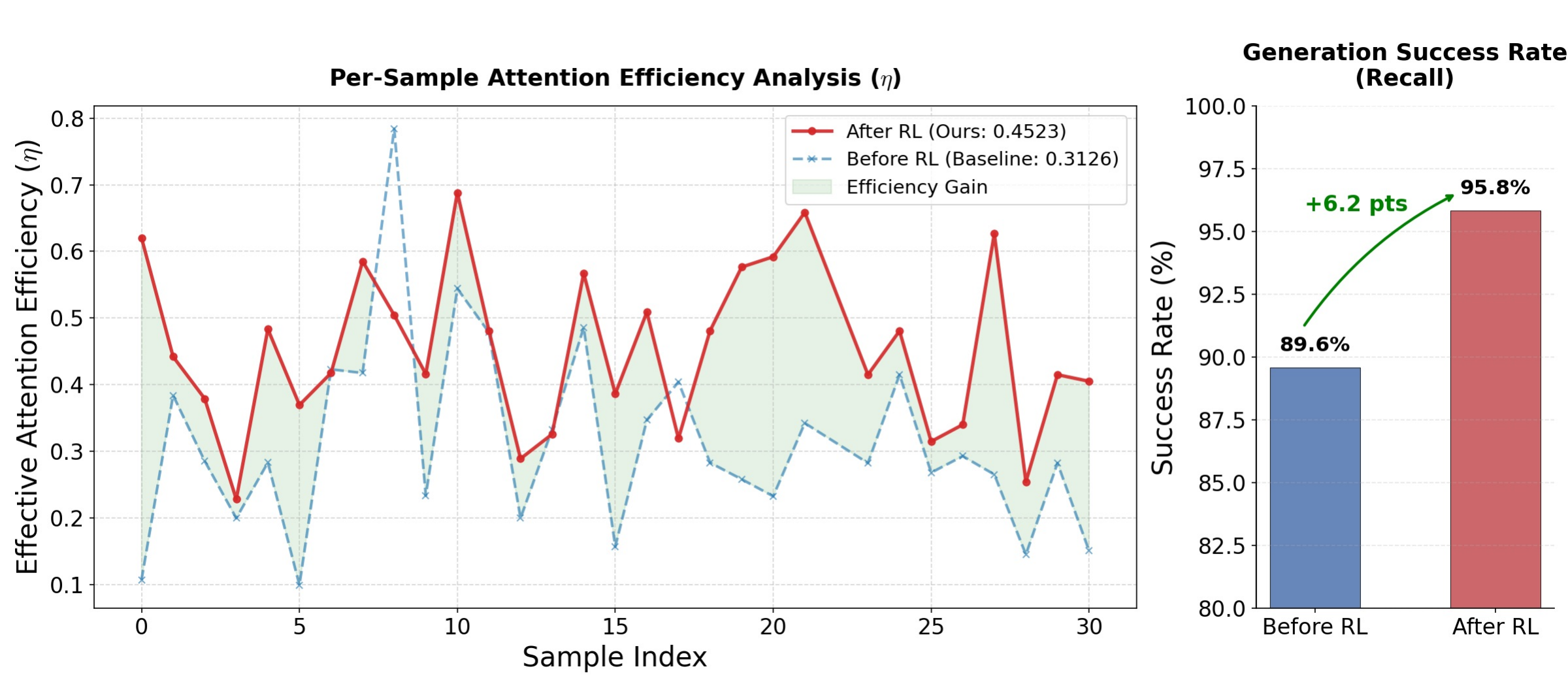}
    \vspace{-4.5mm}
    \caption{\xr{\textbf{Quantitative evaluation on a subset ($30$ samples) of CVTG-Hard.} \textbf{Left}: The \textbf{Effective Attention Efficiency ($\eta$)} analysis. The values displayed in the legend represent the mean $\eta$ scores, which increase from 0.3126 to 0.4523, corresponding to a \textbf{44.7\% relative improvement} in attention quality. \textbf{Right}: The \textbf{Recall} metric increases by 6.2 pts, verifying the method's effectiveness in handling complex character structures.}}
    \label{fig:rl_analysis2}
\end{figure*}

\noindent 1) \xr{\emph{Target Extraction $\&$ Instance Preprocessing.} 
Given a multi-text prompt, we treat each target string $vt_{i}$ as an isolated entity in the ground truth set $VT = \{vt_{1}, vt_{2}, \dots, vt_{n}\}$. Both $VT$ and the OCR-detected~\cite{cui2025paddleocr30technicalreport} results from the generated image are normalized (lowercase conversion and removal of non-alphanumeric characters) to ensure that the insulation performance is measured purely on semantic accuracy rather than formatting noise. }

\noindent 2) \xr{\emph{Isolated Fuzzy Matching.} 
We utilize the Fuzzy Partial Ratio metric to compute an independent similarity score $s_i \in [0, 1]$ for each target $vt_{i}$. Formally, given the global OCR output sequence $\mathcal{O}$ and a target string $vt_{i}$, the partial ratio score is defined as the maximum Levenshtein similarity~\cite{levenshtein1966binary} between $vt_{i}$ and any substring $w$ extracted from $\mathcal{O}$ that has the same length as $vt_{i}$:
\begin{equation}
s_i = \max_{w \subseteq \mathcal{O}, |w| = |vt_{i}|} \left( 1 - \frac{\text{dist}(vt_{i}, w)}{|vt_{i}|} \right),
\end{equation}
where $\text{dist}(\cdot, \cdot)$ denotes the Levenshtein edit distance. 
By matching each target individually against the global OCR output in sliding-window manner, we precisely monitor the ``insulation" quality of each text instance, ensuring that artistic variations or minor recognition errors in one area do not unfairly penalize other text regions.}

\noindent 3) \xr{\emph{Insulation-aware Aggregation.} 
To prevent the model from collapsing by generating one text while neglecting others (insulation failure), we define the base reward $R_{base}$ to balance average performance and individual integrity:
\begin{equation}
R_{base} = (1-\lambda_{bal}) \cdot \frac{1}{n}\sum_{i=1}^{n} s_i + \lambda_{bal} \cdot \min(s_1, \dots, s_n),
\end{equation}
where $\lambda_{bal}$ (empirically set to $0.3$) is a balancing coefficient that controls the sensitivity to the worst-case instance (\emph{i.e.}, the \textbf{bottleneck}). 
The $\min$ term is critical for multi-text insulation, as it explicitly \emph{penalizes the omission or corruption} of any single text target, forcing the model to ``insulate" and preserve all requested instances.}

\noindent 4) \xr{\emph{Anti-interference Penalty.}
A common pathology in reinforcement learning for text generation is the ``text explosion" phenomenon, where the model generates excessive irrelevant text or repetitions to maximize the recall probability (i.e., reward hacking). To mitigate this, we introduce a length-based noise penalty $\lambda_{noise}$. Specifically, we define a tolerance threshold $\delta$ for the length ratio. If the total predicted length $l_{pred}$ exceeds the sum of insulated targets $l_{target}$ by this threshold, the reward is decayed:

\vspace{-1mm}
\begin{equation}
    \lambda_{noise} = \begin{cases} 1
    .0, & \text{if } \frac{l_{pred}}{l_{target}} \le \delta, \\ \frac{1}{(l_{pred}/l_{target}) - \delta+1}, & \text{otherwise.} \end{cases}
\end{equation}
We empirically set $\delta = 1.5$ to accommodate minor OCR redundancies while strictly penalizing excessive gibberish. The final reward is formulated as $R_{ocr} = \lambda_{noise} \cdot R_{base}$. 
In this work, we employ the DiffusionNFT~\cite{zheng2025diffusionnft} to optimize this objective. 
However, it is worth noting that our reward design is algorithm-agnostic and can be seamlessly adapted to other advanced reinforcement learning frameworks~\cite{liu2025flow, xue2025dancegrpo}. 
The proposed penalty ensures the model generates the required text in a clean, insulated manner.} 

\xr{Functionally, these two components establish a dual-constraint mechanism on the generation length. The \emph{Insulation-aware Aggregation} acts as a \textbf{lower bound} constraint, encouraging the model to generate sufficient tokens to cover all targets (preventing omission), while the \emph{Anti-interference Penalty} serves as an \textbf{upper bound} constraint, suppressing the tendency to over-generate (preventing hallucination). This ``push-pull'' dynamic forces the model to converge on the optimal, insulated text layout.}

\xr{To further demonstrate the effectiveness of our insulated RL strategy, we provide qualitative evidence by analyzing the \textbf{aggregated phrase-level} cross-attention maps across different scenes. 
As illustrated in Figure~\ref{fig:rl_analysis}, the baseline model often suffers from feature leakage, where the activation regions of text targets drift to visually salient but unrelated areas (\emph{e.g.}, the attention for `New Year Special' erroneously focusing on the central lantern, or `Fashion' bleeding into surrounding signage). Conversely, after RL fine-tuning, the attention maps for target phrases become notably disentangled and spatially concentrated. This observation corroborates that our reward formulation successfully guides the model to allocate exclusive latent regions for each text target, effectively preventing ``text explosion'' or overlap phenomena at the feature level.}

\begin{figure*}[t!]
  \centering
  \includegraphics[width=0.86\linewidth]{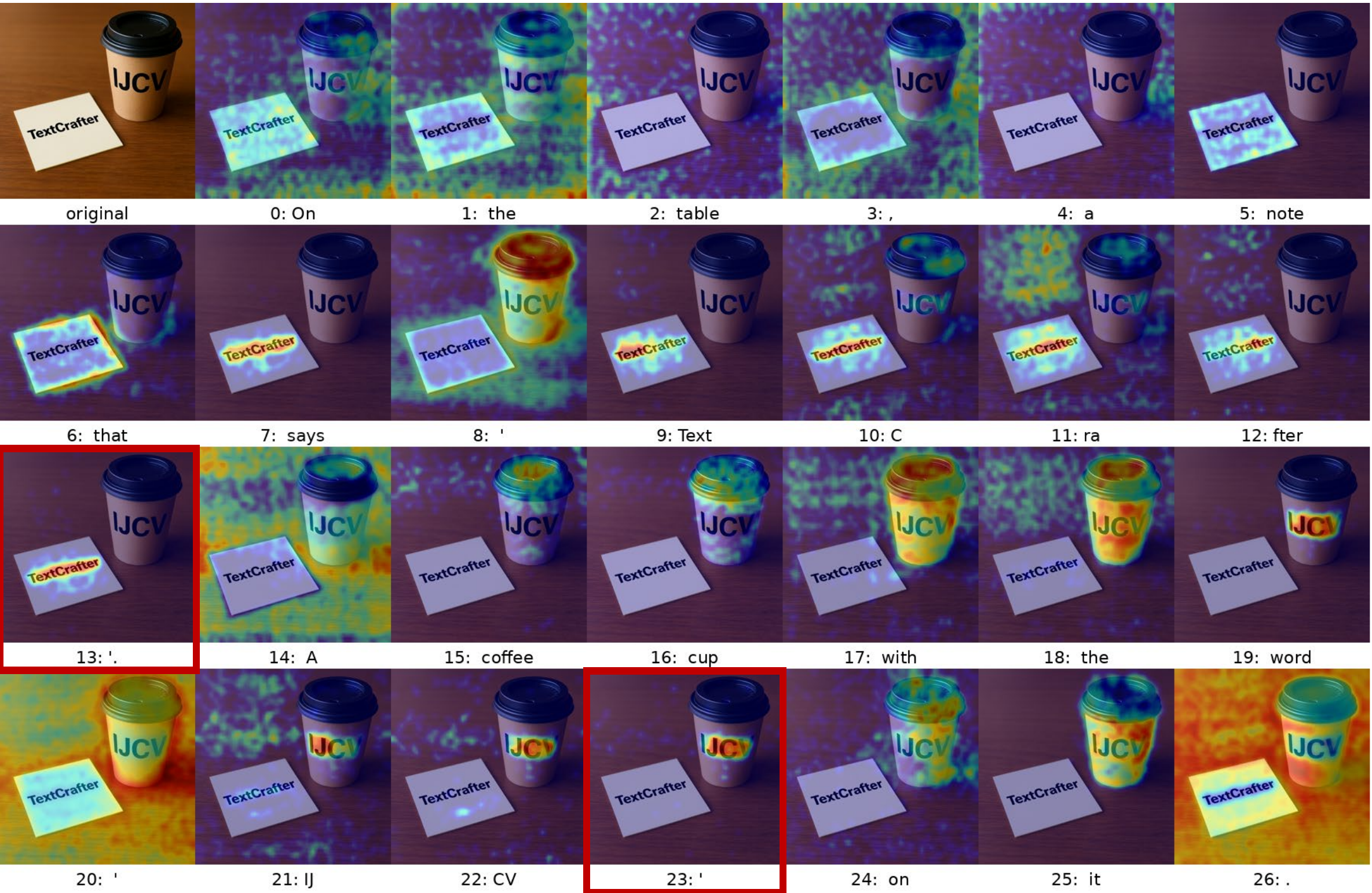}
  \caption{\yt{\textbf{Visualization of prompt tokenization} for ``On the table, a note that says `TextCrafter'. A coffee cup with the word `IJCV' on it.'' along with per-token attention maps.
  In Qwen-Image, each closing quotation mark functions as a spatial anchor, enforcing alignment between text token and its corresponding carrier token and yielding clean spatial disentanglement without cross-text interference.
  }
  }
  \label{fig:qianyinhao}
  \vspace{-2mm}
\end{figure*}

\noindent \xr{
\textbf{Discussion on RL for Multi-Text Generation.} 
We randomly select 30 prompts from the CVTG-Hard and conduct a comparative evaluation between the baseline model and our RL-finetuned model. 
During inference, we selectively extract the cross-attention maps corresponding to the tokens of the target text. 
Subsequently, for each target instance enclosed in quotation marks, we aggregate the maps of its constituent tokens to derive a phrase-level attention map, denoted as $\mathbf{A}^p$.
%
We introduce two metrics to reflect the \textit{omission} and \textit{hallucination} degree, \emph{i.e.} \textbf{Recall} and \textbf{Effective Attention Energy} ($\eta$), respectively.

\noindent\textbf{1) Recall.} 
To quantify the degree of text omission, we calculate the object-level generation success rate. Let $N_{total}$ be the total number of target phrases in the evaluation set, and $N_{succ}$ be the count of generated phrases that are successfully detected and matched with valid bounding boxes. The Recall is defined as:
\begin{equation}
    \text{Recall} = \frac{N_{succ}}{N_{total}} \times 100\%.
\end{equation}
A higher Recall indicates a lower omission rate and robust instruction following.

\begin{figure*}[t!]
    \centering
    \includegraphics[width=0.7\linewidth]
    {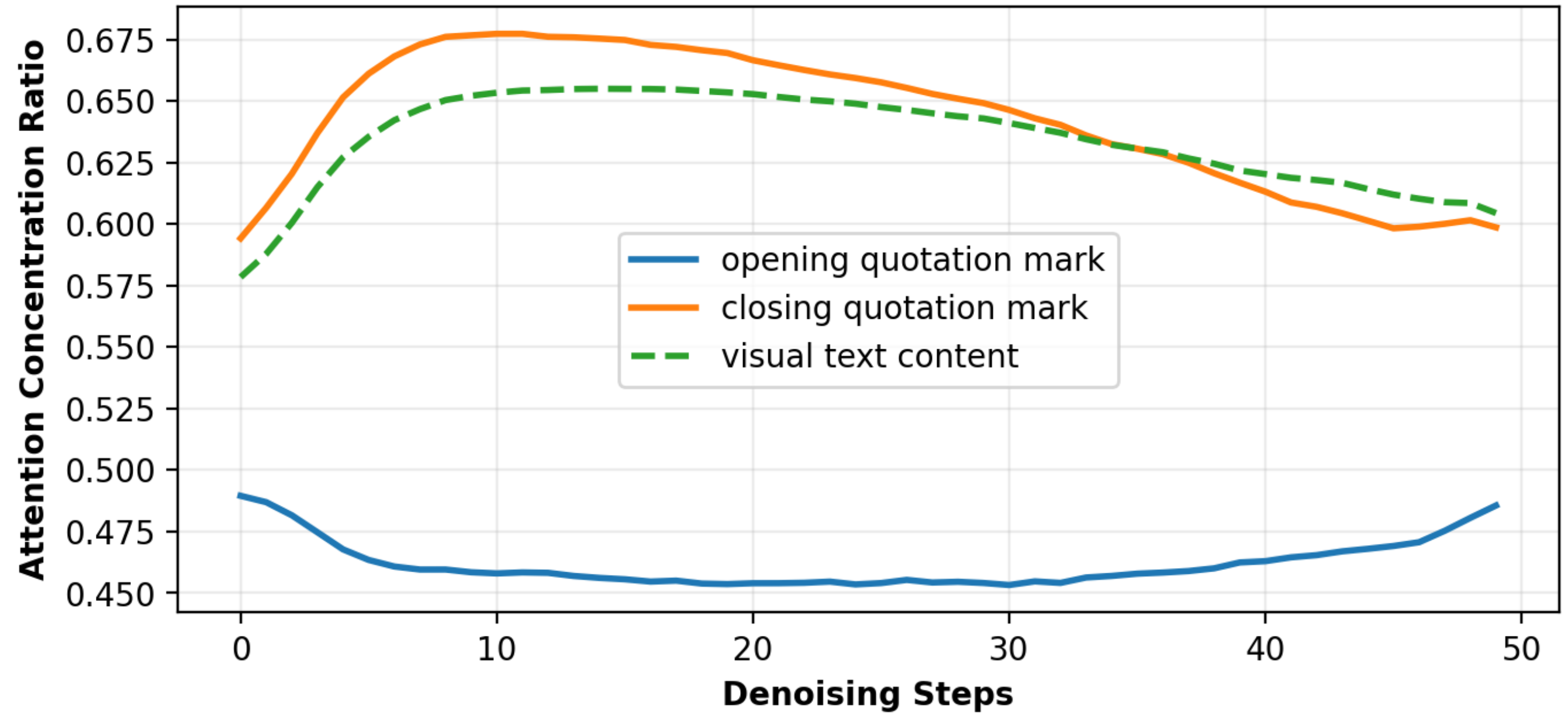}
    \caption{
    \yt{\textbf{Attention concentration ratio over denoising steps in Qwen-Image.}
    Following TextGuider~\cite{baek2025textguider}, we compute the attention concentration ratio as the mean image-to-text cross-modal attention within the OCR-detected text bounding box divided by the global mean over the entire image, and averaged over 100 successful instances.
    The closing quotation mark shows the earliest and strongest attention concentration, acting as a region-level layout anchor. 
    Visual text tokens follow a similar but weaker trend due to finer-grained attention and dilution from loose OCR bounding boxes. 
    }}
    \label{fig:3curve}
\end{figure*}

\vspace{0.5mm}
\noindent\textbf{2) Effective Attention Efficiency ($\eta$).} 
To measure the degree of visual hallucination (\emph{i.e.}, attention leakage), we propose $\eta$ to evaluate the concentration of effective attention energy. 
As illustrated in Figure~\ref{fig:rl_analysis}, we construct the Ground-Truth (GT) phrase-level regions using manually annotated {\color{red}red bounding boxes}. 
The core intuition of this metric is to \emph{maximize the effective attention energy accumulated within the GT box while minimizing the energy leaking into the outside background}. 
Given the phrase-level attention map $\mathbf{A}^p$, we first apply an adaptive statistical threshold to filter out background noise. 
Let $\mu(\mathbf{A}^p)$ and $\sigma(\mathbf{A}^p)$ denote the mean and standard deviation of the map, respectively. We obtain a denoised map $\hat{\mathbf{A}}^p$ by retaining only the signals exceeding the mean plus one standard deviation:
\vspace{-1mm}
\begin{equation}
    \hat{\mathbf{A}}^p_{i,j} = 
    \begin{cases} 
    \mathbf{A}^p_{i,j}, & \text{if } \mathbf{A}^p_{i,j} > \mu(\mathbf{A}^p) + \sigma(\mathbf{A}^p) \\
    0, & \text{otherwise}.
    \end{cases}
\end{equation}
Then, $\eta$ is calculated as the ratio of effective energy inside the target bounding box $\mathcal{R}$ to the energy in the background region:
\begin{equation}
    \eta = \frac{\sum_{(i,j) \in \mathcal{R}} \hat{\mathbf{A}}^p_{i,j}}{\sum_{(i,j) \notin \mathcal{R}} \hat{\mathbf{A}}^p_{i,j} + \xi},
\end{equation}
where $\xi$ is a small constant for numerical stability. 
A higher $\eta$ signifies that the model's high-confidence attention is strictly insulated within the target region, reflecting suppressed hallucinations.
%
The quantitative results are presented in Figure~\ref{fig:rl_analysis2}. 
As shown in the right bar chart, the RL fine-tuning yields a solid improvement in instruction following, increasing the \textbf{Recall} by \textbf{6.2 pts} (from 89.6\% to 95.8\%). 
More importantly, the left line chart visualizes the per-sample attention quality. The \textbf{Effective Attention Efficiency ($\eta$)} demonstrates a substantial \textbf{44.7\% relative lift}. 
This significant boost confirms that our method actively suppresses background leakage, reallocating high-confidence attention to target text regions to mitigate visual hallucinations.

\vspace{1mm}
\noindent \textbf{Post-training Settings}. 
We conducted the RL post-training on {only} $4$ NVIDIA A100 GPUs. 
For detailed hyperparameters, we primarily follow the configuration of Edit-R1~\cite{li2025uniworldv2}. 
Motivated by the curriculum learning strategy~\cite{zheng2025diffusionnft}, we adopt a two-stage training process.
In the first stage, we fine-tune the model on the MixGRPO~\cite{li2025mixgrpounlockingflowbasedgrpo} training dataset for $1,000$ steps, utilizing a weighted combination of PickScore~\cite{kirstain2023pick}, CLIPScore~\cite{hessel2021clipscore}, and HPSv2.1~\cite{wu2023human} rewards to preserve both aesthetic quality and semantic consistency. Subsequently, to specifically optimize text generation, we continue training for another $1,000$ steps with the proposed OCR reward $R_{ocr}$. 
This phase utilizes the English and Chinese text rendering subsets from the Qwen-Image-Self-Generated-Dataset~\cite{qwenimage_dataset}, which collectively comprise approximately $60,000$ prompts, covering diverse scenarios for multi-text rendering.
}

\begin{figure}[t!]
  \centering
  \includegraphics[width=\linewidth]{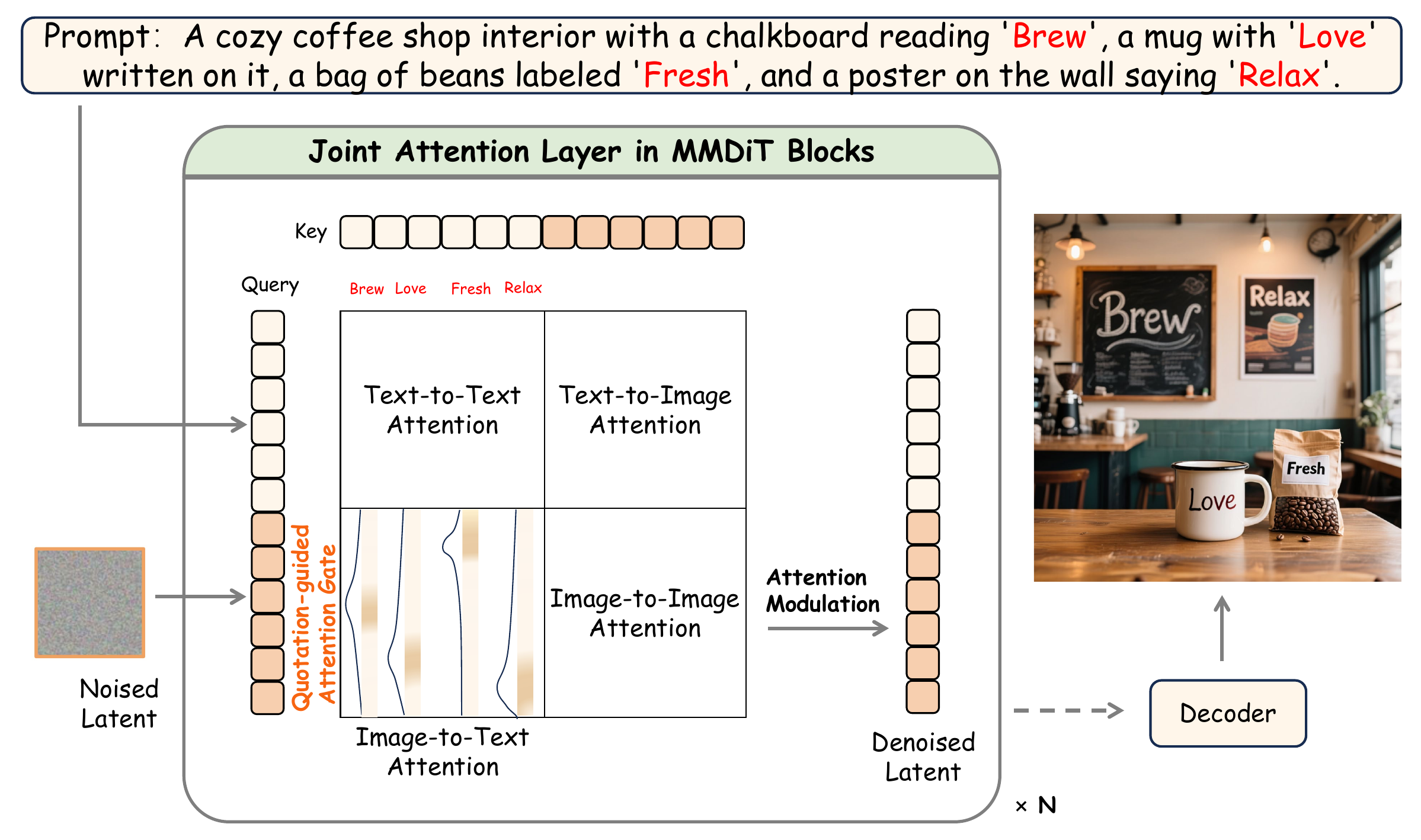}
  \vspace{-2mm}
  \caption{\textbf{Text-oriented Attention with Quotation-guided Attention Gate.} 
  Within the Joint Attention Layer of MMDiT blocks, we leverage closing quotation marks as spatial anchors. 
  The \textbf{Quotation-guided Attention Gate} is constructed from these anchors to dynamically modulate the Image-to-Text attention maps. 
  Applied during inference, this mechanism enforces the concentration of text-related visual tokens within their designated regions, effectively mitigating feature leakage.}
  \label{fig:method_attention}
\end{figure}

\begin{figure*}[t]
  \centering
  \includegraphics[width=0.915\linewidth]{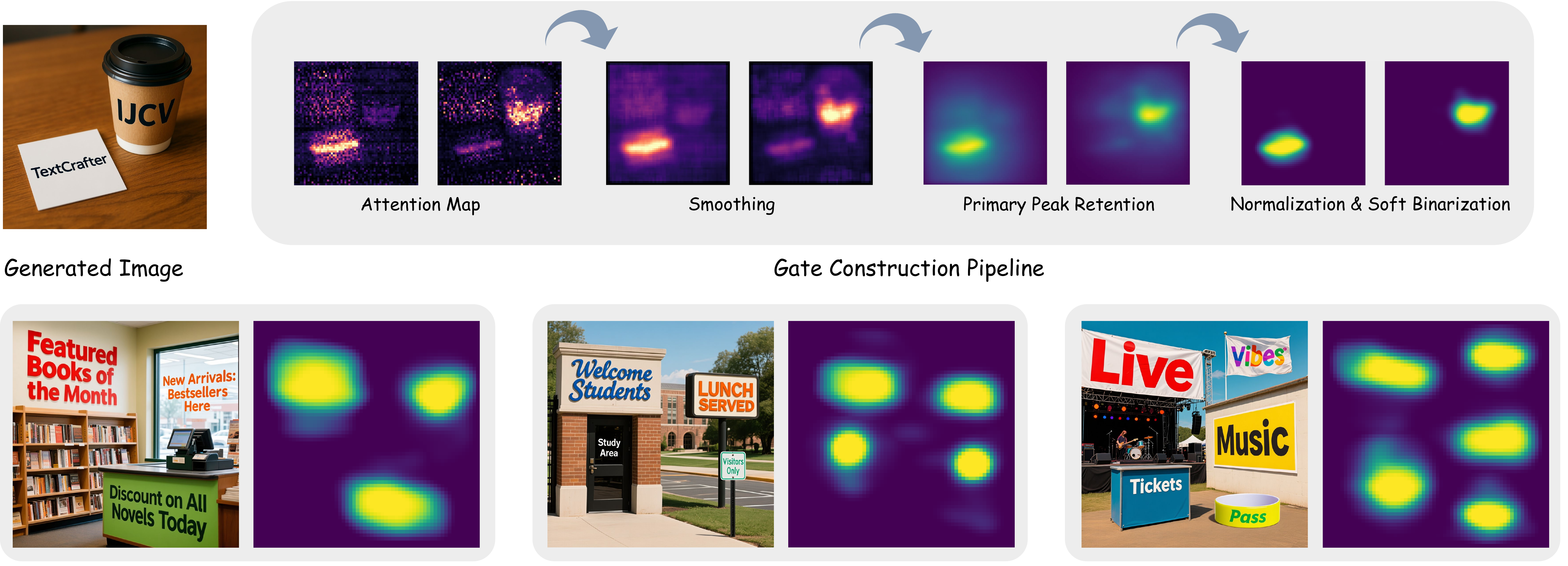}
  \caption{\textbf{Visualization of the Gate Construction Pipeline (Top) and Examples in Complex Scenarios (Bottom).}
  \textbf{Top}: Taking the \emph{anchor quotation marks} of two visual text instances as examples, we observe that directly using raw attention maps as gates is challenging due to inherent noise and multiple high-activation regions.
  To address this issue, we refine the map through three steps:
  \textbf{1) Smoothing} removes scattered noise points in non-target regions.
  \textbf{2) Primary Peak Retention} retains only the single region with the highest activation.
  \textbf{3) Normalization \& Soft Binarization} ensures continuous gate values bounded within $[0, 1]$, where the target region approaches 1 and other regions approach 0.
  \textbf{Bottom}: We provide additional examples in complex scenarios featuring \emph{varying numbers of text instances} (ranging from 3 to 5), demonstrating the robustness and adaptability of the gate mechanism in complex scenarios.
  In each example pair, the left side shows the Generated Image, and the right side displays the visualization of all gates simultaneously.}
  \label{fig:gate} 
\end{figure*}

\subsubsection{\yt{Text-oriented Attention}}\label{sec:attn}
\textbf{Effectiveness of Anchor Quotation Marks.}
It is crucial to prevent the attention of visual text tokens from leaking into the background and ensure their precise concentration within the designated regions.
Rather than modifying the text embeddings directly, we leverage the attention map of the quotation mark as a robust spatial anchor to guide the generation of visual text tokens.
As illustrated in Figure~\ref{fig:qianyinhao}, anchor quotation marks (\emph{i.e.}, closing quotation marks) consistently span the entire rendered text region they govern, suggesting they capture holistic, region-level information about the visual text layout.

We further validate this behavior quantitatively in Figure~\ref{fig:3curve}, following the attention concentration analysis in~\cite{baek2025textguider}.
At denoising step $t$, we extract the image-to-text cross-modal attention $\mathbf{A}^{(t)}=\mathrm{softmax}\!\left(\mathbf{Q}_{\text{img}}\mathbf{K}_{\text{text}}^\top/\sqrt{d}\right)$, and define the attention map of a target text token $\tau$ over image tokens as $a^{(t)}_{\tau}(p)=\mathbf{A}^{(t)}_{p,\tau}$.
Given the OCR-detected text bounding box mask $\mathcal{B}$ (a set of image tokens) and the full image token set $\mathcal{I}$, the {Attention Concentration Ratio} (ACR) is computed as
\vspace{-1mm}
\begin{equation}
    \mathrm{ACR}^{(t)}(\tau)=
    \frac{\frac{1}{|\mathcal{B}|}\sum_{p\in\mathcal{B}} a^{(t)}_{\tau}(p)}
    {\frac{1}{|\mathcal{I}|}\sum_{p\in\mathcal{I}} a^{(t)}_{\tau}(p)}.
    \label{eq:acr}
\end{equation}
\vspace{-1mm}
For the \emph{visual text content} inside the quotation marks, we first average the token-wise attention maps $\bar a^{(t)}_{\text{text}}(p)=\frac{1}{|\mathcal{T}_{\text{text}}|}\sum_{\tau\in\mathcal{T}_{\text{text}}} a^{(t)}_{\tau}(p)$ and then compute $\mathrm{ACR}^{(t)}(\text{text})$ by Eq.~\eqref{eq:acr}.
\yt{As shown in Figure~\ref{fig:3curve}, 
closing quotation marks in Qwen-Image demonstrates the earliest and strongest attention concentration.
Visual text tokens follow a similar but weaker trend due to finer-grained attention and dilution from loose OCR bounding boxes.}
This observation motivates us to utilize the attention map of the dominant quotation mark to construct a spatial gate, which dynamically \emph{highlights the text region and suppresses cross-text interference for visual text tokens}.

\vspace{1mm}
\noindent \textbf{Quotation-guided Attention Gate.}
We propose a \emph{Text-oriented Attention} mechanism (see Figure~\ref{fig:method_attention}) that employs a 
quotation-guided gate to modulate image-to-text attention.
Let $vt_k$ denote the $k$-th visual text instance. 
Let $q_k$ be its corresponding anchor quotation mark, and $\mathcal{C}_k$ be the set of visual text tokens within the quotation marks.
At step $t$, we utilize attention map from previous step $t-1$ to construct a spatial gate $G_k^{(t)}$, which enhances the attention of visual text tokens in $\mathcal{C}_k$.

\begin{figure*}[t!]
  \centering
  \includegraphics[width=0.72\linewidth]{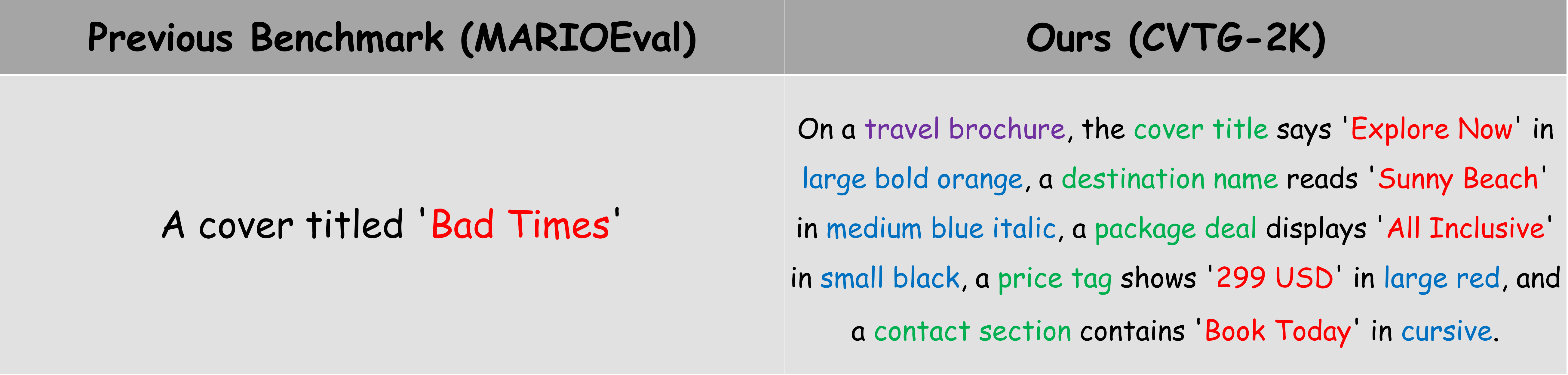}
  \caption{\textbf{Comparison of prompt complexity and granularity.} 
  \textbf{Left (MARIOEval~\cite{chen2024textdiffuser})}: Prompts typically contain only a single text content with minimal context. 
  \textbf{Right (CVTG-2K)}: Our prompts feature global context descriptions ({\color{mypurple}Purple}), multiple text content ({\color{red}Red}), specific position requirements ({\color{mygreen}Green}), and fine-grained visual attributes ({\color{blue}Blue}), providing a more challenging and realistic evaluation for generative models.}
  \label{fig:benchmark}
\end{figure*}

\noindent 1) \textit{Gate Construction.}
First, we extract the cross-attention map of the anchor quotation mark $q_k$, averaged over all $L$ layers and $H$ heads, indexed by $\ell$ and $h$ respectively. 
Let $p$ denote the spatial index. The raw map is $\bar{a}^{(t-1)}_{k}(p) = \frac{1}{L \cdot H} \sum_{\ell, h} \mathbf{A}^{(t-1, \ell)}_{h, p, q_k}$.
To generate a robust gate, we apply a specific operator $\mathcal{G}$ (see Figure~\ref{fig:gate}), consisting of smoothing, primary peak retention, and normalization \& soft binarization.
\begin{itemize}
    \vspace{-2mm}
    \item \textbf{Smoothing:} We apply a $5$$\times$$5$ average pooling to smooth out  scattered noise points: $\tilde{a}_k = \operatorname{AvgPool}(\bar{a}^{(t-1)}_k)$.
    
    \item \textbf{Primary Peak Retention:} We identify the peak position $p^*_k = (x^*_k, y^*_k) = \arg\max_p \tilde{a}_k(p)$ and suppress secondary peaks using a Gaussian proximity mask. 
    The Gaussian widths $\sigma_{x,k}, \sigma_{y,k}$ are adaptively determined by the second central moments of smoothed attention map $\tilde{a}_k$, representing the spatial extent around the peak:
    \begin{equation}
    \vspace{-3mm}
    \begin{split}
        \sigma_{x,k} &= \sqrt{\frac{\sum_{p}\tilde a_k(p)(x-x^*_k)^2}{\sum_{p}\tilde a_k(p)}}, \\
        \sigma_{y,k} &= \sqrt{\frac{\sum_{p}\tilde a_k(p)(y-y^*_k)^2}{\sum_{p}\tilde a_k(p)}},
    \end{split}
    \vspace{-2mm}
    \end{equation}
    where $(\sigma_{x,k}, \sigma_{y,k})$ critically determine the spatial extent of the gate, thereby controlling whether the gate is overly concentrated or overly diffuse. 
    The region-masked map is then computed as $r_k(p) = \tilde{a}_k(p) \cdot \exp(-\frac{(x-x^*_k)^2}{2\sigma_{x,k}^2} - \frac{(y-y^*_k)^2}{2\sigma_{y,k}^2})$.

   \item \textbf{Normalization \& Soft Binarization:} To form the final gate $G_k^{(t)}(p) \in [0, 1]$, we first apply max-norm normalization: $\hat{r}_k(p) = r_k(p) / \max_p r_k(p)$. 
    Then, we define dynamic thresholds $v_{low} = \operatorname{Quantile}_{0.8}(\hat{r}_k)$ and $v_{high} = \operatorname{Quantile}_{0.99}(\hat{r}_k)$. The gate is formulated as:
    \begin{equation}
        G_k^{(t)}(p) = \operatorname{smoothstep}\left( \frac{\hat{r}_k(p) - v_{low}}{v_{high} - v_{low}} \right),
    \end{equation}
    where $\operatorname{smoothstep}(z) = z^2(3-2z)$ maps values to $[0, 1]$.
\end{itemize}

\noindent 2) \textit{Attention Modulation.}
The constructed gate $G_k^{(t)}$ indicates the precise region where the $k$-th text should appear. We enhance the attention of all visual text tokens $\tau \in \mathcal{C}_k$ strictly within this gated region.
%
%
Finally, the attention map is updated as:
\begin{equation}
    \mathbf{A}^{(t)}_{p, \tau} \leftarrow \mathbf{A}^{(t)}_{p, \tau} \cdot \left[ 1 + G_k^{(t)}(p) 
    \right], \quad \forall \tau \in \mathcal{C}_k.
\end{equation}
This mechanism ensures that visual text tokens focus intensely on the text region defined by the anchor quotation mark, significantly mitigating text misgeneration and blurriness without disrupting the global layout.

\begin{figure*}[t!]
    \centering
    \includegraphics[width=0.95\linewidth]{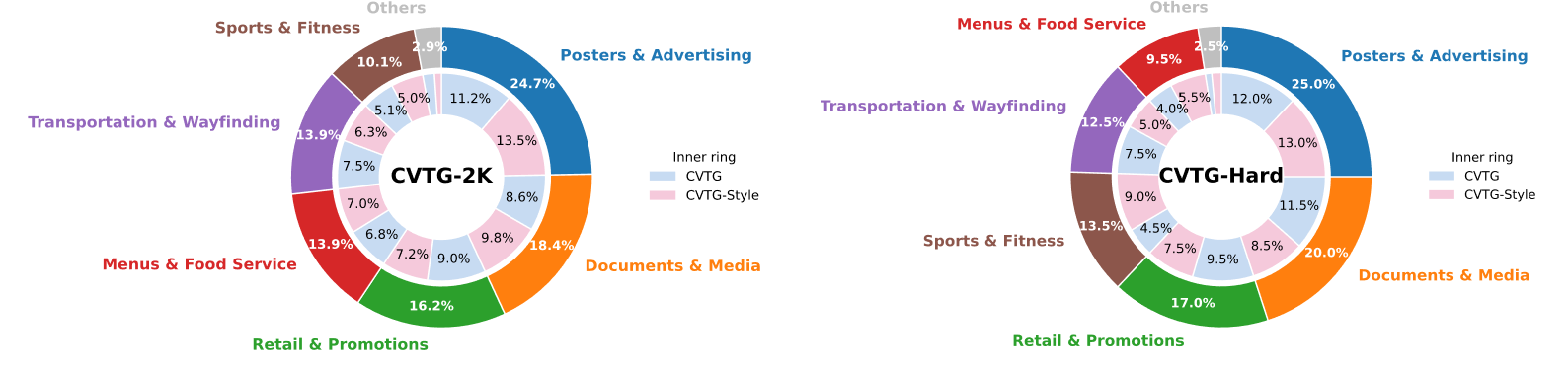}
    \caption{\textbf{\yt{Scene distributions of CVTG-2K (Left) and CVTG-Hard (Right).}}
    Outer ring shows coarse category frequencies, and inner ring indicates the split between CVTG and CVTG-Style.}
    \label{fig:scene}
\end{figure*}

\subsection{CVTG Benchmark}\label{sec:Benchmark}
\yt{Currently, there is no publicly available benchmark dataset specifically designed for complex visual text generation. 
To address this gap, we introduce CVTG-2K, a challenging public benchmark for complex visual text generation.
All prompts are generated through the OpenAI's O1-mini API~\cite{OpenAI_O1Mini}, which encompass a variety of scenes containing complex visual texts. }
Unlike previous datasets synthesized using fixed rules, CVTG-2K ensures the diversity and rationality of the data distribution. 
We first prompt the O1 model to conceive a scene and then imagine the possible visual texts that may appear in that scene, utilizing Chain-of-Thought (CoT)~\cite{wei2022chain} to ensure the quality of the generated prompts. 
\yt{Further details can be found in the supplementary material.} 
Through rigorous filtering rules and meticulous post-processing, we have generated 2,000 prompts containing complex visual texts. 
On average, the visual texts in CVTG-2K contain 8.10 words and 39.47 characters, surpassing all previously published visual text benchmark datasets in terms of visual text length. 
Table~\ref{tab:benchmark} presents a comparison of the data in CVTG-2K with previous benchmark datasets. 
Furthermore, 
CVTG-2K is the \emph{first} benchmark dataset to incorporate multiple visual text regions, with the number of regions ranging from 2 to 5. 
The distribution across region counts is approximately 20\%, 30\%, 30\%, and 20\%, respectively.
With diverse numbers of text regions and varying visual text lengths, CVTG-2K enables comprehensive evaluation of model performance on complex visual text generation. Following a careful review process, we ensure CVTG-2K contains no discriminatory or inflammatory content.


\yt{To further enhance the challenge of CVTG-2K, we partition the benchmark into two subsets: \textbf{CVTG} and \textbf{CVTG-Style}. 
\textbf{CVTG-Style} contains half of the data, where each visual text is randomly augmented with natural-language attribute annotations, while the remaining half in \textbf{CVTG} has no attributes. 
The attributes include size (large, medium, small), color, and font (regular, bold, italic, cursive). 
Each text in \textbf{CVTG-Style} is assigned one or more attributes, facilitating systematic evaluation of visual text stylization and customization.
Additionally, we provide fine-grained information.
We employ O1-mini to decompose each prompt into descriptions \( D = \{d_1, d_2, \dots, d_n\} \) containing multiple visual texts, and to concisely express the correspondence between each text instance and its position using critical words, which serve as the carriers of the visual text.
Meticulous manual review ensured the accuracy of the fine-grained information.
}

\vspace{1mm}
\noindent \textbf{Scene Distributions of CVTG-2K.}
\yt{
Figure~\ref{fig:benchmark} contrasts prompt complexity and granularity between prior benchmark MARIOEval and CVTG-2K.
Figure~\ref{fig:scene} (Left) shows the scene distribution of CVTG-2K.
The distribution is obtained by assigning each prompt to one coarse category using a local Qwen3-30B-A3B-Instruct-2507~\cite{qwen3} classifier with a predefined taxonomy and aggregating category frequencies.
CVTG-2K has been publicly released together with the code and adopted to evaluate the visual text generation capabilities of multiple academic and industrial text-to-image models, including Qwen-Image~\cite{wu2025qwen}, Z-image~\cite{cai2025z}, LongCat-Image~\cite{team2025longcat}, EMU3.5~\cite{cui2025emu3}, GLM-Image~\cite{glmimage} and Ovis-Image~\cite{wang2025ovis}, contributing to research on text rendering.}

\vspace{1mm}
\noindent \textbf{{Construction of CVTG-Hard.}}
\zn{
We further introduce CVTG-Hard, a challenging subset comprising the most difficult prompts from CVTG-2K, together with their Chinese translations, yielding 400 test samples.
%
As shown in Table~\ref{tab:benchmark}, the visual texts in CVTG-hard contain an average of 8.61 words and 40.79 characters. Figure~\ref{fig:scene} (Right) shows CVTG-Hard still maintains the diverse scene distribution.
}

\vspace{-3mm}
\yt{
\section{Experiments}
\subsection{Evaluation and Baselines}
\noindent \textbf{Evaluation Metrics and Datasets.}
The experiments employ five metrics: Word Accuracy and Normalized Edit Distance (NED)~\cite{tuo2023anytext} are employed to evaluate the accuracy of text rendering. CLIPScore~\cite{hessel2021clipscore}, VQAScore~\cite{lin2024evaluating}, and Aesthetics~\cite{LAION2022Aesthetics} are utilized to assess the quality of the generated images. 
For detailed configurations of the metrics calculation, please refer to the supplementary materials.
%
We evaluate TextCrafter on four datasets: The proposed CVTG-2K, CVTG-Hard, LongText-Bench~\cite{geng2025x} and Geneval~\cite{ghosh2023geneval}.
Unlike previous benchmarks that primarily focus on short text or single-scene scenarios, LongText-Bench is specifically designed to evaluate the accuracy of generating extended text content in both English and Chinese across diverse scenarios. 
}

\noindent \textbf{Baselines.} 
\yt{
We compare TextCrafter with extensive state-of-the-art Text-to-Image models.
Representative baselines from academia include AnyText~\cite{tuo2023anytext}, TextDiffuser2~\cite{chen2025textdiffuser}, RAG-Diffusion~\cite{chen2025region}, 3DIS~\cite{zhou20243dis}, and Stable Diffusion 3.5 Large~\cite{esser2024scaling}.
%
%
For a comprehensive comparison, we additionally compare TextCrafter against a wide array of models/products from the industry, including GPT Image~\cite{gptimage}, Gemini-2.5~\cite{comanici2025gemini}, Seedream~\cite{seedream2025seedream}, Qwen-Image~\cite{wu2025qwen}, HunyuanImage-3.0~\cite{cao2025hunyuanimage}, Longcat-Image~\cite{team2025longcat}, Z-Image~\cite{cai2025z}, EMU3.5~\cite{cui2025emu3}, GLM-Image~\cite{glmimage}, FLUX.1~\cite{flux}. 
This setup allows us to rigorously validate TextCrafter's effectiveness in complex visual text generation tasks.
}

\begin{table*}[t]
  \centering
  \caption{\textbf{Quantitative results on CVTG-2K dataset}. 
  TextCrafter is compared against state-of-the-art models from both industry and academia.
  `*' denotes results cited from the previous papers.
  \textbf{Bold} values denote the best performance, while \underline{underlined} values indicate the second-best performance for each metric.
  The three parenthesized values in the last row (\emph{e.g.}, +13.4\%) denote the relative gains over the baseline model Qwen-Image on the corresponding metrics.
  }
  \label{tab:CVTG-2K}
  \footnotesize
  \resizebox{\textwidth}{!}{%
  \begin{tabular}{lccccc}
    \toprule
  Model & Word Accuracy ($\uparrow$) & NED ($\uparrow$) & CLIPScore ($\uparrow$) & VQAScore ($\uparrow$) & Aesthetics ($\uparrow$) \\ 
    \midrule
    FLUX.1 dev {\color{gray}\tiny{[Black Forest Labs 2024]}} & 0.4965 & 0.6879 & 0.7401 & 0.8886 & \underline{5.91} \\
    GPT Image 1 [High] {\color{gray}\tiny{[OpenAI 2025]}}* & 0.8569 & 0.9478 & 0.7982 & - & -  \\
    Gemini 2.5 Flash Image {\color{gray}\tiny{[Google 2025]}}* & 0.7364 & 0.8516 & - & - & - \\
    Seedream 4.5 {\color{gray}\tiny{[ByteDance 2025]}}* & 0.8990 & 0.9483 & 0.8069 & - & - \\
    Qwen-Image {\color{gray}\tiny{[Alibaba 2025]}}* & 0.8288 & 0.9116 & 0.8017 & - & - \\
    Z-Image {\color{gray}\tiny{[Alibaba 2025]}}* & 0.8671 & 0.9367 & 0.7969 & - & - \\
    HunyuanImage-3.0 {\color{gray}\tiny{[Tencent 2025]}}* & 0.7650 & 0.8765 & \underline{0.8121} & - & - \\
    Longcat-Image {\color{gray}\tiny{[Meituan 2025]}}* & 0.8658 & 0.9361 & 0.7859 & - & - \\
    Emu3.5 {\color{gray}\tiny{[BAAI 2025]}}* & \underline{0.9123} & \underline{0.9656} & - & - & - \\
    GLM-Image {\color{gray}\tiny{[Z.ai 2026]}}* & {0.9116} & {0.9557} & 0.7877 & - & - \\
    \midrule
    SD3.5 Large~{\color{gray}\tiny{[ICML 2024]}} & 0.6548 & 0.8470 & 0.7797 & \underline{0.9297} & 5.56 \\
    AnyText~{\color{gray}\tiny{[ICLR 2024]}} & 0.1804 & 0.4675 & 0.7432 & 0.6935 & 4.53 \\
    TextDiffuser-2~{\color{gray}\tiny{[ECCV 2024]}} & 0.2326 & 0.4353 & 0.6765 & 0.5627 & 4.51 \\
    RAG-Diffusion~{\color{gray}\tiny{[ICCV 2025]}} & 0.2648 & 0.4498 & 0.6688 & 0.6397 & 5.58 \\
    3DIS~{\color{gray}\tiny{[ICLR 2025]}} & 0.3813 & 0.6505 & 0.7767 & 0.8684 & 4.86 \\
    \midrule
    \textbf{TextCrafter (Qwen-Image)} & \textbf{0.9400}$^{(+13.4\%)}$ & \textbf{0.9757}$^{(+7.0\%)}$ & \textbf{0.8305}$^{(+3.6\%)}$ & \textbf{0.9570} & \textbf{5.90} \\
    \bottomrule
  \end{tabular}}
\end{table*}

\begin{table*}[]
\caption{\textbf{Quantitative results on CVTG-Hard dataset}.
For ZH, due to the absence of explicit whitespace word boundaries in Chinese, evaluation is performed at the span level rather than the word level. 
To accommodate spans split across lines, we allow matches formed by concatenating consecutive OCR lines. 
\textbf{Span Accuracy} is the fraction of ground-truth spans exactly matched by such concatenations.
NED is the average, over all spans, of the normalized edit distance between each ground-truth span and its best-matching line concatenation.
}

\label{tab:CVTG-2K-Hard}
\centering
\tiny
\begin{tabular}{lcccc}
    \toprule
    \multirow{2}{*}{\textbf{Model}}     & \multicolumn{2}{c}{\textbf{EN}}      & \multicolumn{2}{c}{\textbf{ZH}}       \\ 
    \cmidrule(lr){2-3} \cmidrule(lr){4-5}
     & Word Accuracy ($\uparrow$) & NED ($\uparrow$) & Span Accuracy ($\uparrow$) & NED ($\uparrow$) \\ 
    \midrule
    FLUX.1 dev~{\color{gray}\tiny{[Black Forest Labs 2024]}}                            & 0.2427                 & 0.4612       & 0.0000                 & 0.0104       \\
    SD3.5~{\color{gray}\tiny{[ICML 2024]}}                              & 0.4623                 & 0.7078       & 0.0014                 & 0.0105       \\
    Qwen-Image~{\color{gray}\tiny{[Alibaba 2025]}}                         & 0.6312                 & 0.7776       & 0.6526                 & 0.8237       \\
    Z-Image~{\color{gray}\tiny{[Alibaba 2025]}}                            &    0.7218	& 0.8477	& 0.7125	& 0.8548            \\
    Longcat-Image~{\color{gray}\tiny{[Meituan 2025]}}                      &     0.7991	& 0.8919 & 0.6894 & 0.8415  \\
    HunyuanImage3.0~{\color{gray}\tiny{[Tencent 2025]}}                 & 0.6719	& 0.8221 & 0.5821 & 0.7315    \\
    GLM-Image~{\color{gray}\tiny{[Z.ai 2026]}}                          &  \underline{0.8171}	& \underline{0.9000}   & \underline{0.8610}	& \underline{0.9164}   \\
    \midrule
    \textbf{TextCrafter (Qwen-Image)} & \textbf{0.8862}$^{(+40.4\%)}$ & \textbf{0.9470}$^{(+21.8\%)}$ & \textbf{0.8692}$^{(+33.2\%)}$ & \textbf{0.9518}$^{(+15.6\%)}$ \\
    \bottomrule
\end{tabular}
\end{table*}

\begin{figure*}[t!]
  \centering
  \includegraphics[width=\linewidth]{figures/Longtext-bench.png}
  \vspace{-3mm}
  \caption{\textbf{Quantitative comparisons on LongText-Bench.} 
  We report the \textbf{Text Accuracy} on English (EN), Chinese (ZH), and their Average (Avg). 
  Only models from industry (\emph{e.g.}, Qwen-Image, GPT Image 1 [High], Z-image \emph{etc.}) are used for comparison, as existing academic approaches~\cite{tuo2023anytext,chen2025region,zhou20243dis,chen2025textdiffuser} exhibit substantially inferior performance, often achieving scores below 0.5.
  }
  \label{tab:LongTextBench}
\end{figure*}

\begin{table*}
\caption{\textbf{Quantitative results on Geneval}.   
All results are cited from the previous papers.
}
\label{tab:Geneval}
\centering
\resizebox{\textwidth}{!}{
\begin{tabular}{lccccccc}
\toprule
Model                            & \multicolumn{1}{l}{Single Object($\uparrow$)} & \multicolumn{1}{l}{Two Object($\uparrow$)} & \multicolumn{1}{l}{Counting($\uparrow$)} & \multicolumn{1}{l}{Colors($\uparrow$)} & \multicolumn{1}{l}{Position($\uparrow$)} & \multicolumn{1}{l}{Attribute Binding($\uparrow$)} & \multicolumn{1}{l}{Overall($\uparrow$)} \\ \hline
FLUX.1 dev {\color{gray}\tiny{[Black Forest Labs 2024]}}                & 0.98                              & 0.81                           & 0.74                         & 0.79                       & 0.22                         & 0.45                                  & 0.66                        \\
GPT Image 1 [High] {\color{gray}\tiny{[OpenAI 2025]}}           & \underline{0.99}                              & 0.92                           & 0.85                         & 0.92                       & \underline{0.75}                         & 0.61                                  & 0.84                        \\
Seedream 4.0 {\color{gray}\tiny{[ByteDance 2025]}}                     & \underline{0.99}                              & 0.92                           & 0.72                         & 0.91                       & \textbf{0.76}                         & 0.74                                  & 0.84                        \\
Qwen-Image {\color{gray}\tiny{[Alibaba 2025]}}                      & \underline{0.99}                              & 0.92                           & \underline{0.89}                         & 0.88                       & \textbf{0.76}                         & \underline{0.77}                                  & \underline{0.87}                        \\
Longcat-Image {\color{gray}\tiny{[Meituan 2025]}}                    & \underline{0.99}                              & \textbf{0.98}                           & 0.86                         & 0.86                       & \underline{0.75}                         & 0.73                                  & \underline{0.87}                        \\
Z-Image {\color{gray}\tiny{[Alibaba 2025]}}                         & \textbf{1.00}                                 & 0.95                           & 0.78                         & \textbf{0.93}                       & 0.62                         & \underline{0.77}                                  & 0.84                        \\ \hline
Show-o {\color{gray}\tiny{[ICLR 2025]}}                 & 0.95                              & 0.52                           & 0.49                         & 0.82                       & 0.11                         & 0.28                                  & 0.53                        \\
PixArt-$\alpha$ {\color{gray}\tiny{[ICLR 2024]}}               & 0.98                              & 0.50                            & 0.44                         & 0.80                        & 0.08                         & 0.07                                  & 0.48                        \\
SD3.5 Large {\color{gray}\tiny{[ICML 2024]}}           & 0.98                              & 0.89                           & 0.73                         & 0.83                       & 0.34                         & 0.47                                  & 0.71                        \\
Lumina-Image 2.0 {\color{gray}\tiny{[ICCV 2025]}}       & -                                 & 0.87                           & 0.67                         & -                          & -                            & 0.62                                  & 0.73                        \\ \hline
\textbf{TextCrafter(Qwen-Image)} & \underline{0.99}                              & \underline{0.97}                           & \textbf{0.90}                          & \underline{0.92}                       & 0.73                         & \textbf{0.83}                                  & \textbf{0.88}                        \\ 
\bottomrule
\end{tabular}
}
\end{table*}

%

\vspace{-2mm}
\subsection{Quantitative Results}

\noindent \textbf{Comparison on CVTG-2K.}
As demonstrated in Table~\ref{tab:CVTG-2K}, TextCrafter surpasses competing methods in both text accuracy and image quality on CVTG-2K, with TextCrafter~(Qwen-Image) improving word accuracy by 13.4\% relative to Qwen-Image. 
While Stable Diffusion 3.5 performs adequately in simple scenarios, its efficacy diminishes significantly with increased textual complexity. AnyText and TextDiffuser-2, despite being trained on rule-based data, fail to generalize to multi-region tasks. Similarly, RAG-Diffusion and 3DIS struggles with visual text generation despite its multi-instance capabilities. 
By comparison, industry models~\cite{wu2025qwen,cui2025emu3,seedream2025seedream} achieve significantly greater performance improvements than those from academia.
While TextCrafter is primarily tailored for complex multi-text generation, it nevertheless exhibits exceptional robustness in rendering extended single-text sequences. 

\begin{figure*}[t]
  \centering
  \includegraphics[width=\linewidth]{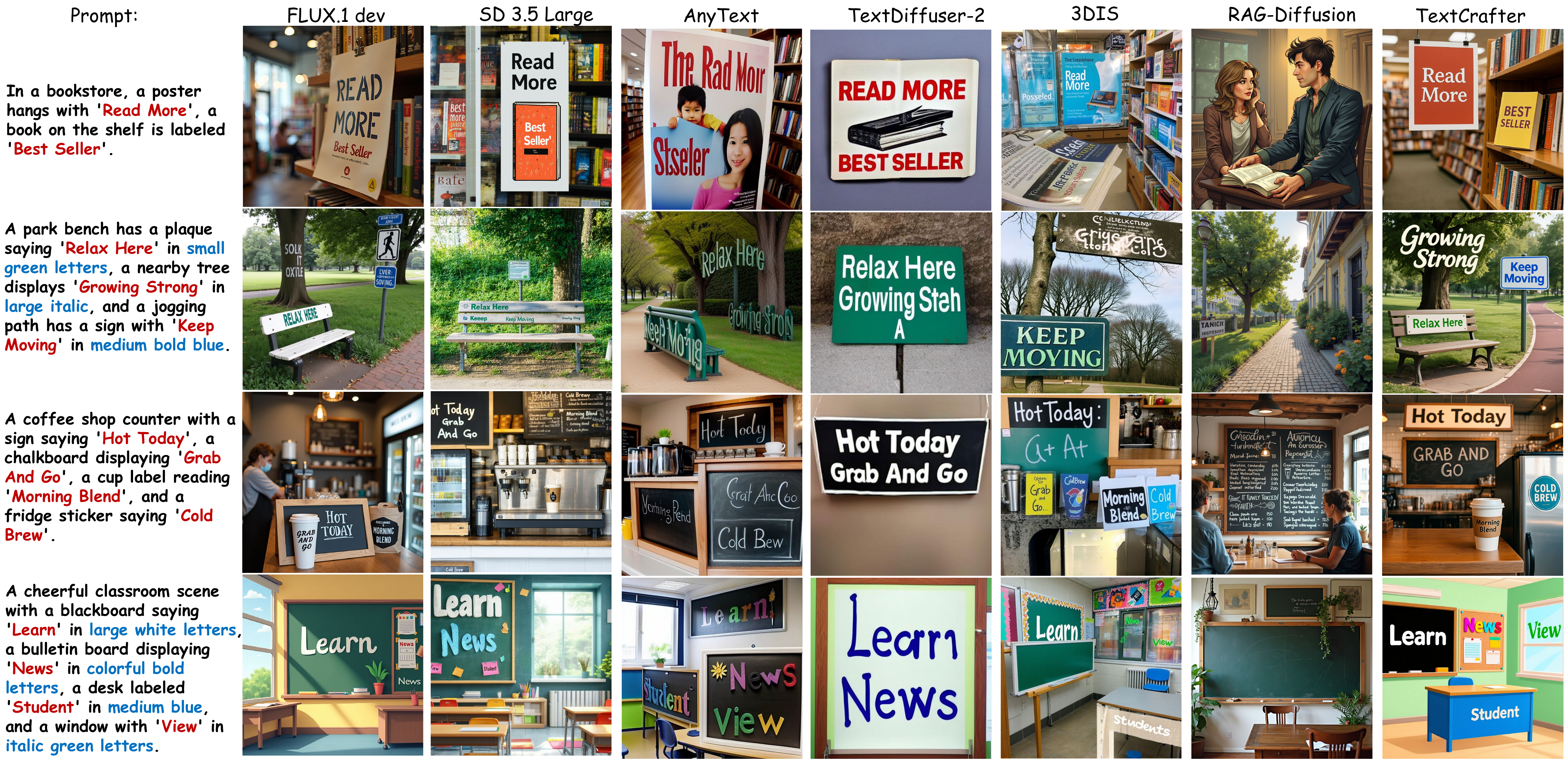}
  \vspace{-2mm}
  \caption{\textbf{Qualitative comparison of TextCrafter with other baselines on CVTG-2K}.
  In the prompts, red content denotes the required visual text, while blue content denotes the attributes of visual text.
  TextCrafter excels in delivering harmonious and aesthetically pleasing images. It also accurately renders multiple visual texts while maintaining stability in complex scenarios.}
  \label{fig:Qualitative1}
\end{figure*}

\begin{figure*}[t]
  \centering
  \includegraphics[width=0.995\linewidth]{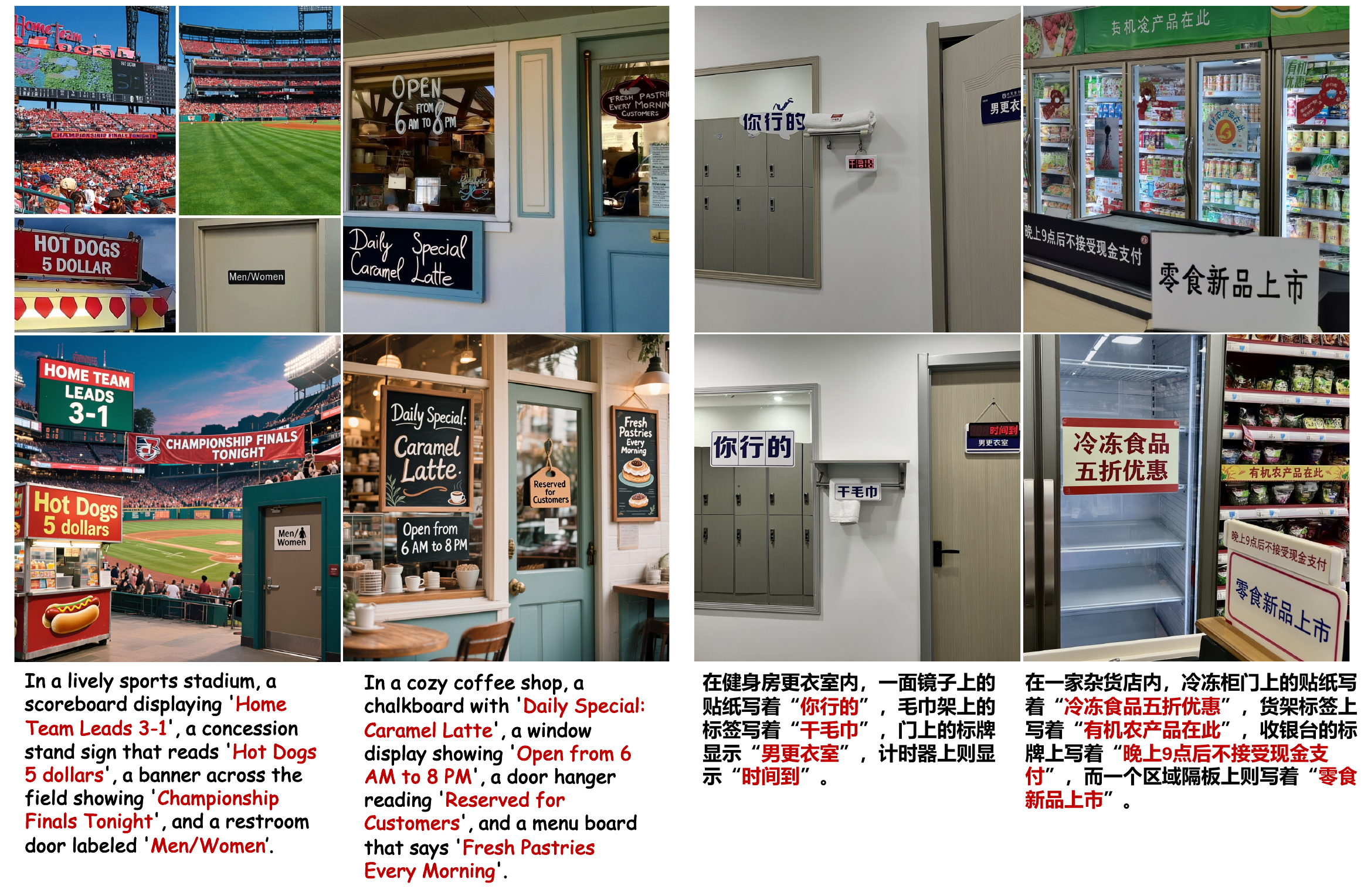}
  
  \caption{\textbf{Qualitative comparison with Qwen-Image on CVTG-Hard (Top: Qwen-Image; Bottom: Ours)}.
    \zn{
      Red contents in the prompts denote the required visual texts.
      }
    }
  \label{fig:CVTG-Hard_Qualitative_Result}
\end{figure*}

\begin{figure*}[t]
  \centering
  \includegraphics[width=\linewidth]{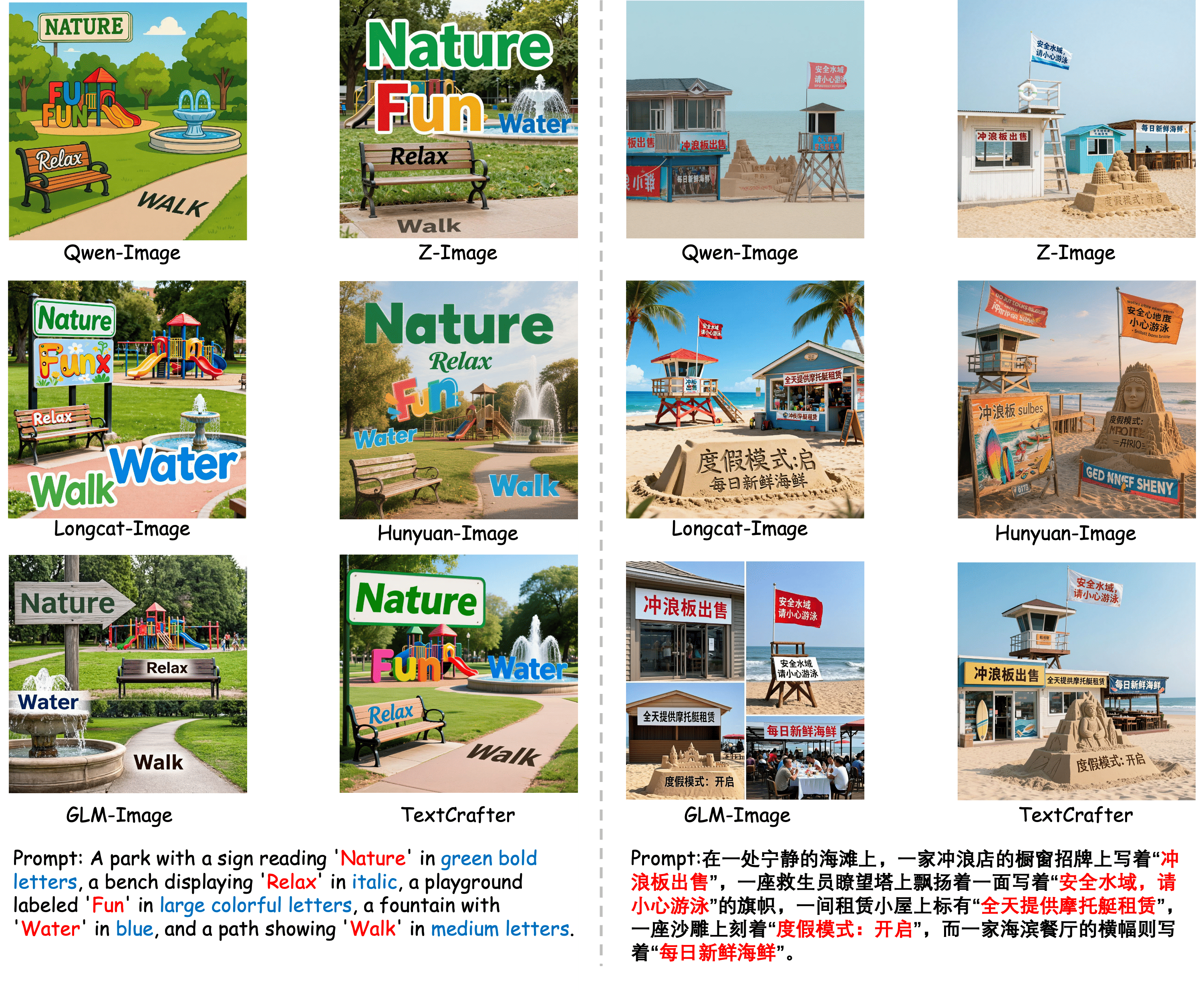}
  \vspace{-5mm}
  \caption{\textbf{Visual comparison on CVTG-Hard against state-of-the-art industrial models}. 
  In the prompt, \textcolor{red}{red} indicates the target visual text, and \textcolor{blue}{blue} indicates the required attributes.
    }
  \label{fig:CVTG-Hard_more}
\end{figure*}

\begin{figure*}[t]
  \centering
  \includegraphics[width=\linewidth]{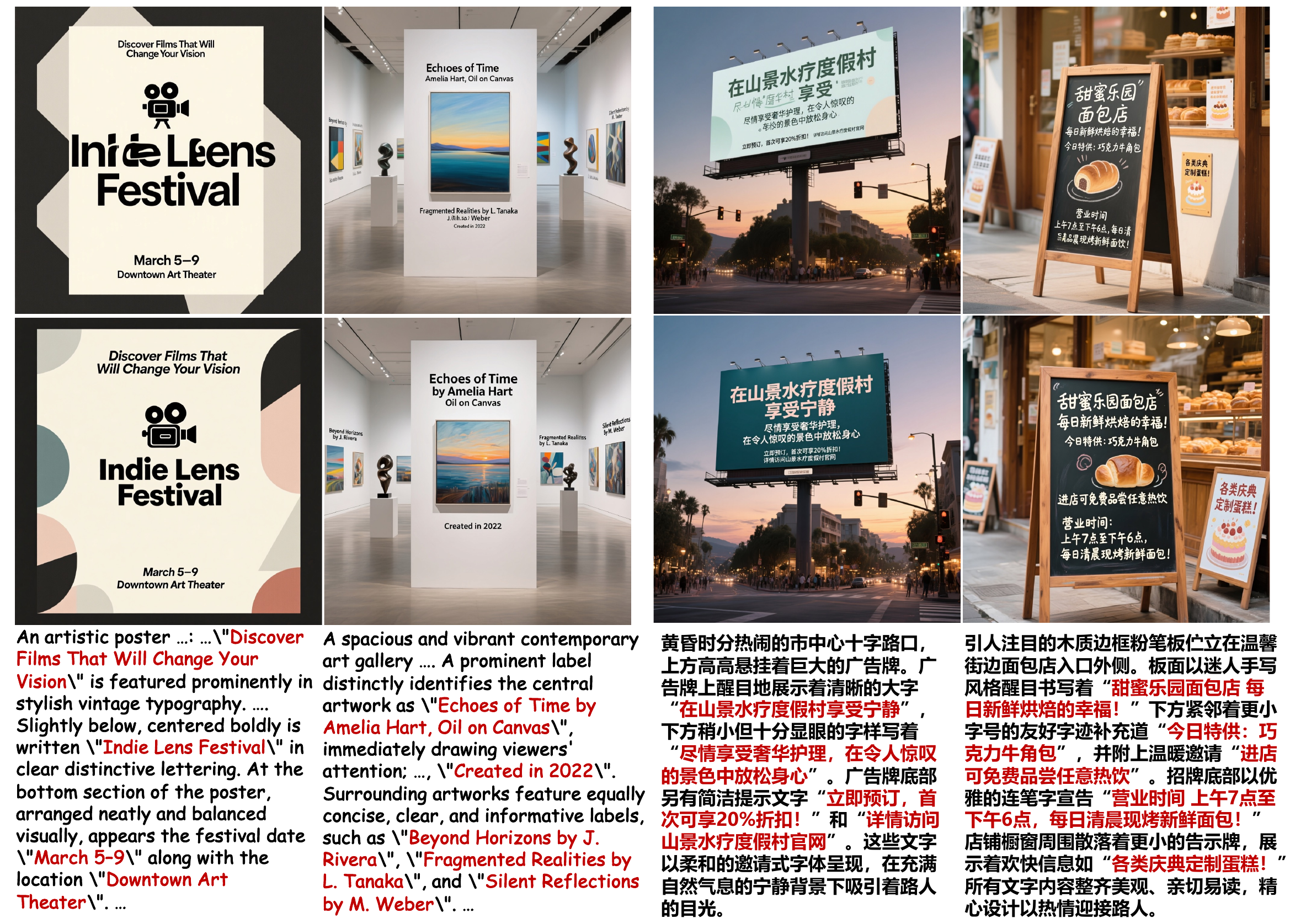}
  
  \caption{\textbf{Qualitative comparison with Qwen-Image on LongText-Bench (Top: Qwen-Image; Bottom: Ours)}.
    \zn{
      Red contents in the prompts denote the required visual texts.
      }
    }
  \label{fig:Longtext_Qualitative_Result_2}
  \vspace{-2mm}
\end{figure*}

\vspace{1.5mm}
\noindent \textbf{Comparison on CVTG-Hard.}
\zn{
We further evaluated TextCrafter on the CVTG-Hard dataset, comparing with state-of-the-art open-source models. 
Note that EMU3.5 is not included, as its inference is relatively time-consuming and our computational resources are limited.
As illustrated in Table~\ref{tab:CVTG-2K-Hard}, the performance of competitive baselines, such as Z-Image and Qwen-Image, declines sharply on this challenging benchmark. Conversely, our model demonstrates exceptional robustness, with its Word/Span Accuracy surpassing 
Qwen-Image by 40.4\% and 33.2\% in English and Chinese scenarios, respectively.
}

\noindent \textbf{Comparison on LongText-Bench.}
\yt{
As shown in Figure~\ref{tab:LongTextBench}, TextCrafter (Qwen-Image) also achieves state-of-the-art performance on LongText-Bench~\cite{geng2025x}, outperforming leading commercial systems (\emph{e.g.}, GPT Image~\cite{gptimage}, Seedream~\cite{seedream2025seedream}) and strong open-source competitors (\emph{e.g.}, Qwen-Image~\cite{wu2025qwen}, EMU3.5~\cite{cui2025emu3}, GLM-Image~\cite{glmimage}). 
Notably, our model secures the competitive text accuracy across both English and Chinese subsets, demonstrating its versatility and precision in handling long texts.
}

\noindent \textbf{Comparison on Geneval.}
\yt{
We additionally evaluated the TextCrafter (Qwen-Image) model on a general-purpose text-to-image benchmark \textbf{Geneval}~\cite{ghosh2023geneval} in Table~\ref{tab:Geneval} to assess its performance in generic scenarios. 
TextCrafter achieved an overall score of 0.88, slightly outperforming the baseline model Qwen-Image (0.87). 
This demonstrates that our approach \emph{significantly enhances complex visual text generation} while maintaining strong performance in general text-to-image tasks.
}

\vspace{-3mm}
\subsection{Qualitative Results}
\noindent \textbf{Visualizations on CVTG-2K Dataset.}
Figure~\ref{fig:Qualitative1} illustrates comparative visual results between TextCrafter and several state-of-the-art academic approaches on CVTG-2K. 
While SD3.5~\cite{esser2024scaling} and FLUX.1 dev~\cite{flux} generate visually appealing images, they exhibit deficiencies in text rendering as regional complexity increases. 
AnyText~\cite{tuo2023anytext} demonstrates suboptimal performance with multi-word text instances, TextDiffuser-2~\cite{chen2025textdiffuser} compromises background fidelity (resulting in diminished CLIPScore), and 3DIS~\cite{zhou20243dis} deteriorates textual information during layout-to-depth conversion. 
While RAG-Diffusion~\cite{chen2025region} excels at region-aware generation, it tends to neglect the rendering of visual text. 
Conversely, TextCrafter achieves superior visual harmony while accurately rendering multiple texts without attribute confusion or prompt deviation.


\vspace{0.5mm}
\noindent \textbf{Visualizations on CVTG-Hard.} 
Figure~\ref{fig:CVTG-Hard_Qualitative_Result} presents the qualitative comparison on the CVTG-Hard dataset, which demands precise text generation across multiple distinct regions. We compare our model with the baseline Qwen-Image. When dealing with challenging tasks involving larger volumes, Qwen-Image frequently exhibits significant issues such as text rendering errors, omissions, and hallucinations. In contrast, our model handles such cases effectively. 
Figure~\ref{fig:CVTG-Hard_more} illustrates additional visual
comparison between TextCrafter and other industrial text-to-image models, including Z-Image, Longcat-Image, HunyuanImage, and GLM-Image.
TextCrafter consistently outperforms these strong industrial models, particularly in \emph{text accuracy} and adherence to \emph{specified attributes}.
More visualizations are provided in the supplementary material.

\vspace{0.5mm}
\noindent \textbf{Visualizations on LongText-Bench.} Figure~\ref{fig:Longtext_Qualitative_Result_2} illustrates the visual results on LongText-Bench, a dataset characterized by high-density textual content. In these scenarios, Qwen-Image is prone to generation degradation, characterized by severe character omissions, spelling inconsistencies, and hallucinatory content as sequence length increases. Conversely, our model maintains robust character integrity.

\begin{table*}[t]
  \centering
  \footnotesize
  \caption{\textbf{Ablation studies} on CVTG-Hard (English) subset with Qwen-Image.}
  \label{tab:ablation}
  \setlength{\tabcolsep}{12pt}
  \begin{tabular}{lcc}
    \toprule
    \textbf{Method} & \makecell{Word  Accuracy ($\uparrow$)} & NED ($\uparrow$) \\
    \midrule
    Baseline (Qwen-Image) & 0.6312 & 0.7776 \\
    \midrule
    + Text Insulation  & 0.8792$^{(+39.3\%)}$  & 0.9369$^{(+20.5\%)}$ \\
    + Text-oriented Attention & 0.7422$^{(+17.6\%)}$ & 0.8598$^{(+10.6\%)}$ \\
    \textbf{+ Both (Text Insulation and Attention)} & \textbf{0.8862}$^{(+40.4\%)}$ & \textbf{0.9470}$^{(+21.8\%)}$ \\
    \bottomrule
  \end{tabular}
\end{table*}

\begin{figure*}[t]
    \centering
    \includegraphics[width=\textwidth]{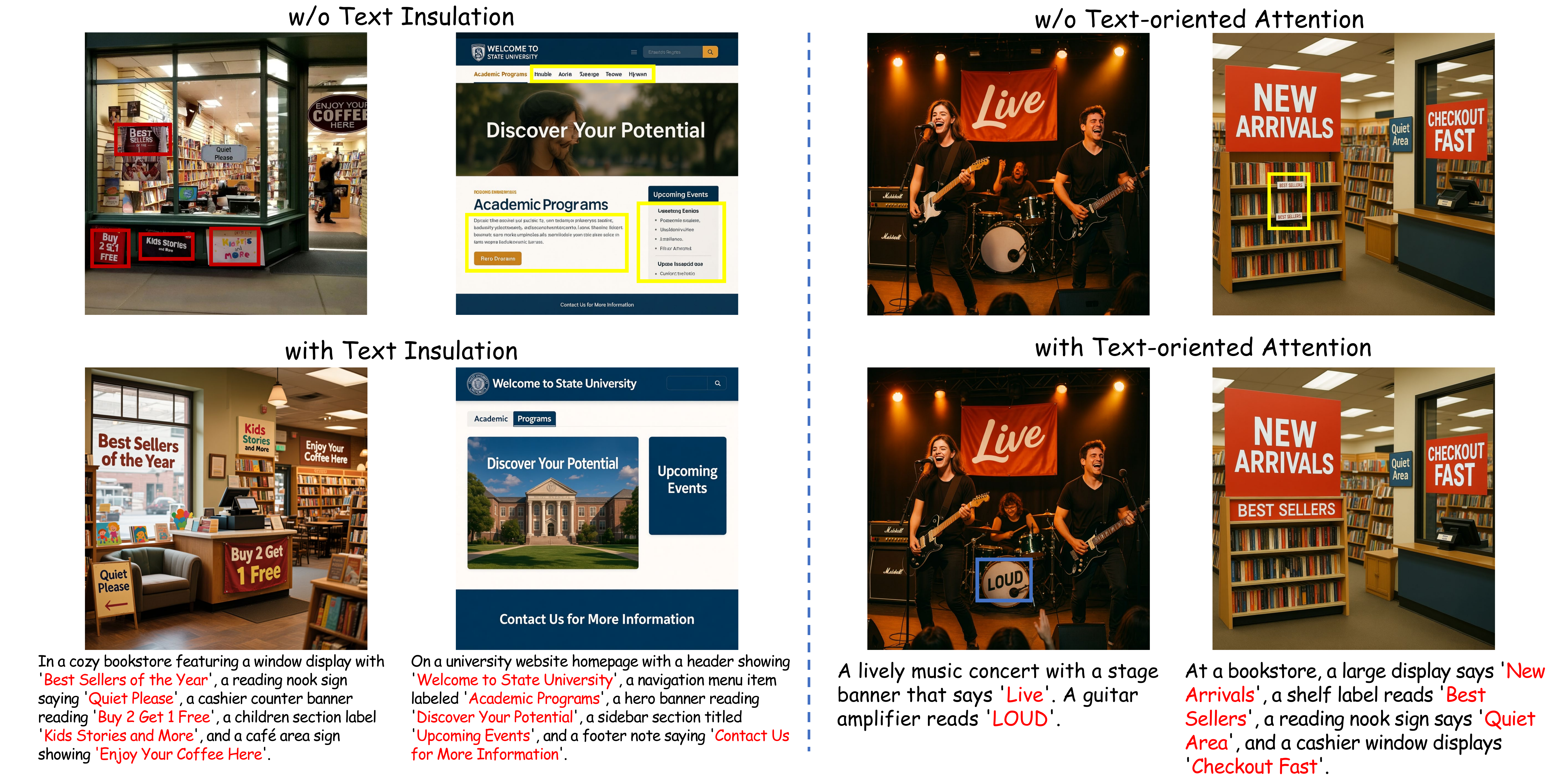}
    \vspace{-3mm}
    \caption{\textbf{Qualitative visualization of ablation studies using Qwen-Image as the baseline model.} 
    Red boxes indicate the occurrence of \textbf{Text Misgeneration}. Yellow boxes represent the occurrence of \textbf{Text Hallucination}, including unintelligible gibberish and repeated text. Blue boxes highlight the resolution of \textbf{Text Omission}.}
    \label{fig:xiaorong}
\end{figure*}

\subsection{Ablation Study}
To validate the effectiveness of the proposed components, we conduct ablation studies on the CVTG-Hard (English) subset. Table~\ref{tab:ablation} presents the quantitative results, while Figure~\ref{fig:xiaorong} provides the qualitative comparisons.

\noindent \textbf{Ablation on Text Insulation.} 
We first evaluate the impact of Text Insulation. 
As evidenced in Table~\ref{tab:ablation}, enabling Text Insulation significantly boosts performance, 
the proposed RL-based insulation raises Qwen-Image from 0.6312 to 0.8792 on Word Accuracy metric.
This improvement stems from the effective decoupling of complex visual texts, effectively mitigating cross-text interference.
Qualitatively, as shown in the top rows of Figure~\ref{fig:xiaorong}, the baseline model frequently suffers from severe degradation. 
Specifically, the first column displays {Text Misgeneration} (red boxes), for example the target ``Buy 2 Get 1 Free'' is erroneously rendered as ``Buy 251'' and ``FIEE''. 
Meanwhile, the second column exhibits extensive {unintelligible gibberish} (yellow boxes) that was not requested in the prompt. 
By enforcing independent generation regions, Text Insulation not only guarantees textual correctness but also effectively eliminates these hallucinated artifacts, yielding clean images with sharp, legible text.

\noindent \textbf{Ablation on Text-oriented Attention.}
This strategy effectively alleviates the \textit{Text Omission} challenge by ensuring concentration on all requested texts.
Quantitative results in Table~\ref{tab:ablation} show that this module further enhances the metrics. 
When combined with Text Insulation (the ``+ Both'' setting), the Word Accuracy and NED peak at 0.8862 and 0.9470, respectively.
Visual comparisons in the bottom rows of Figure~\ref{fig:xiaorong} demonstrate its specific utility. 
Without this module, texts in peripheral or small regions (e.g., the ``LOUD'') are often omitted. 
Enabling Text-oriented Attention successfully recovers these missing texts (highlighted in blue boxes).
Furthermore, the \textit{Primary Peak Retention} strategy employed during the gate construction ensures that attention is concentrated within a single connected region. This spatial constraint effectively suppresses the \textit{text repetition} hallucination, preventing the model from generating duplicate text artifacts.

\begin{table*}[t!]
  \centering
  \caption{\textbf{Effectiveness of the proposed reward function across different RL algorithms.} Evaluated on the CVTG-Hard (English) subset. Our reward model consistently improves performance regardless of the optimization strategy.}
  \label{tab:rl_methods_comparison}
  \footnotesize
  \setlength{\tabcolsep}{10pt}
  \begin{tabular}{lcc}
    \toprule
    \textbf{Method} & Word Acc. ($\uparrow$) & NED ($\uparrow$) \\
    \midrule
    Baseline (Qwen-Image) & 0.6312 & 0.7776 \\ 
    \hline
    \noalign{\vskip 2pt}
    DiffusionNFT (\textbf{with $R_{ocr}$}) & \textbf{0.8792}$^{(+39.3\%)}$ & \textbf{0.9369}$^{(+20.5\%)}$ \\ 
    Flow-GRPO (\textbf{with $R_{ocr}$}) & 0.8699$^{(+37.8\%)}$ & 0.9332$^{(+20.0\%)}$ \\ 
    \bottomrule
  \end{tabular}
\end{table*}

\begin{figure*}
    \centering
    \includegraphics[width=\textwidth]{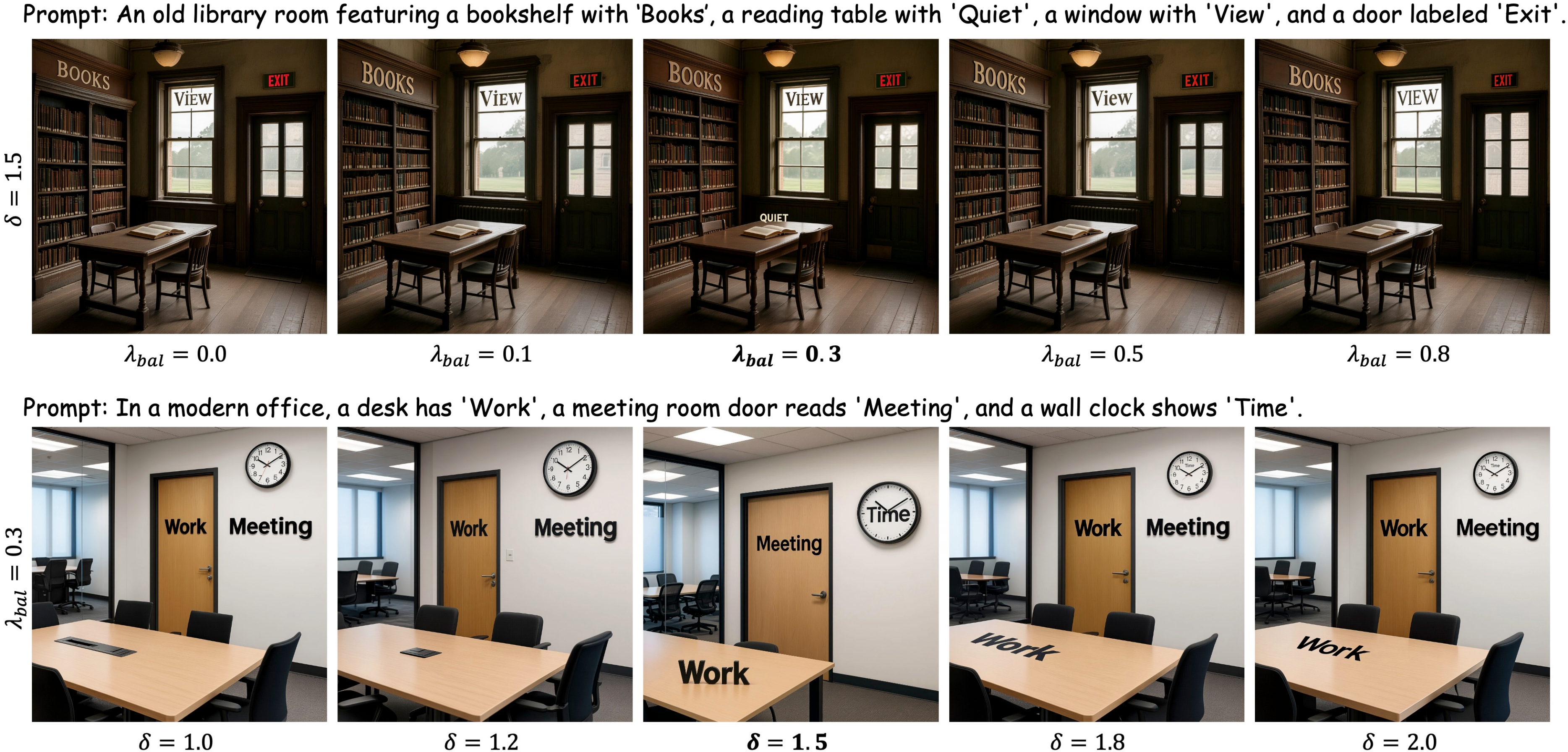}
    
    \caption{
    \textbf{Visual ablation study on the balancing coefficient $\lambda_{bal}$ and length tolerance threshold $\delta$}.
    The top row demonstrates the effect of $\lambda_{bal}$ (with fixed $\delta=1.5$). The difficult text instance ``QUIET'' is omitted at both extremes ($\lambda_{bal}=0.0$ and $0.8$), while the optimal setting ($\lambda_{bal}=0.3$) successfully renders it.
    The bottom row illustrates the impact of $\delta$ (with fixed $\lambda_{bal}=0.3$). Strict thresholds ($\delta \le 1.2$) lead to over-suppression (missing ``Time'' on the clock), whereas loose thresholds ($\delta \ge 1.8$) induce hallucinations (redundant ``Work'' on the table).
    The central column represents our default setting, achieving the best trade-off between completeness and cleanliness.
    }
    \label{fig:rl_ablation}
\end{figure*}

\vspace{1mm}
\noindent \textbf{Ablation on RL Algorithms.} 
We further ablate our reward function $R_{ocr}$ using two RL algorithms: DiffusionNFT and Flow-GRPO. 
As shown in Table~\ref{tab:rl_methods_comparison}, both methods yield substantial improvements over the baseline (Word Acc. 0.6312). Specifically, DiffusionNFT achieves a Word Accuracy of 0.8792 (+39.3\%), while Flow-GRPO attains a comparable 0.8699 (+37.8\%). 
The consistent performance gains across these disparate optimization frameworks demonstrate that our proposed reward model is \textit{robust and algorithm-agnostic, effectively guiding the optimization process regardless of the underlying RL framework.}

\noindent \textbf{Ablation on the Hyperparameters $\lambda_{bal}$ and $\delta$ in Reward Model.}
We conduct ablation studies on two critical hyperparameters: the balancing coefficient $\lambda_{bal}$, which is designed to mitigate omissions by emphasizing worst-case instances, and the length tolerance threshold $\delta$, which serves as a constraint to suppress hallucinations. 
Table~\ref{tab:ablation_RLhyperparams} reveals an inverted U-shape trend for both parameters. $\lambda_{bal}=0.3$ achieves peak accuracy (0.8792) by effectively recovering hard instances, whereas deviations lead to neglect or instability. Similarly, $\delta=1.5$ proves optimal; strict thresholds penalize valid noise, while looser constraints ($\delta=1.8$) cause a performance drop to 0.8554 due to redundancy.
Figure~\ref{fig:rl_ablation} visually confirms these findings. The difficult word ``QUIET'' is omitted at both $\lambda_{bal}$ extremes (due to average-based optimization or instability) but correctly rendered at 0.3. For length tolerance, strict values ($\delta \le 1.2$) cause over-suppression (missing ``Time'' on the clock), while loose values ($\delta \ge 1.8$) induce hallucinations (redundant ``Work'' on the table).

\noindent \textbf{Ablation on Smoothing Kernel Size.}
The smoothing operation is pivotal for mitigating high-frequency noise in raw attention maps. 
We evaluate the impact of the average pooling kernel size $k \times k$ on the CVTG-Hard (English) subset, testing standard sizes $k \in \{3, 5, 7\}$. 
As presented in Table~\ref{tab:pooling_ablation}, the $5 \times 5$ kernel yields the optimal performance. 
Consequently, we adopt a $5 \times 5$ kernel for all experiments.

\begin{table}[t!]
  \centering
  \caption{\textbf{Ablation on hyperparameters $\lambda_{bal}$ and $\delta$} evaluated on the CVTG-Hard (English) subset. \textbf{Bold} denotes the best performance.}
  \label{tab:ablation_RLhyperparams}
  
  \footnotesize
  \setlength{\tabcolsep}{10pt}
  \begin{tabular}{cccc}
    \toprule
    \textbf{$\lambda_{bal}$} & \textbf{$\delta$} & Word Acc. ($\uparrow$) & NED ($\uparrow$) \\
    \midrule
    0.0 & \multirow{5}{*}{1.5} &  0.8776 	& 0.9362  \\
    0.1 &                      &  0.8770 	& 0.9366 \\
    0.3 &             &  \textbf{0.8792}	& \textbf{0.9369}  \\ 
    0.5 &                      & 0.8630     & 0.9325 \\
    0.8 &                      & 0.8788     & 0.9358 \\
    \midrule
    \multirow{4}{*}{0.3} & 1.0 & 0.8734 	& 0.9333  \\
                         & 1.2 & 0.8595 	   & 0.9266  \\
                         & 1.8 & 0.8554 & 0.9243 \\
                         & 2.0 & 0.8702 & 0.9337 \\
    \bottomrule
  \end{tabular}
\end{table}

\begin{table}[t!]
  \vspace{-3mm}
  \centering
  \caption{\textbf{Ablation on Smoothing Kernel Size} in Text-oriented Attention module on CVTG-Hard (English) subset. \textbf{Bold} denotes the best results.}
  \label{tab:pooling_ablation}
  \footnotesize
  \setlength{\tabcolsep}{10pt}
  \begin{tabular}{ccc}
    \toprule
    Kernel Size & Word Acc. ($\uparrow$) & NED ($\uparrow$) \\
    \midrule
    Baseline (Qwen-Image) & 0.6312 & 0.7776 \\
    \midrule
    $3 \times 3$ & 0.7288 & 0.8504 \\
    $5 \times 5$ & \textbf{0.7422} & \textbf{0.8598} \\
    $7 \times 7$ & 0.7369 & 0.8563 \\
    \bottomrule
  \end{tabular}
\end{table}

\begin{figure*}[t!]
    \centering
    \pgfplotsset{
        width=\linewidth,
        height=4.5cm,
        every axis/.append style={
            xlabel={Gaussian Width},
            xmin=5.5, xmax=22.5, 
            xtick={6,8,10,12,14,16,18,20,22}, 
            legend style={at={(0.97,0.03)}, anchor=south east, font=\tiny, fill=none, draw=none},
            label style={font=\footnotesize},
            tick label style={font=\scriptsize},
            grid=major,
            grid style={dashed, gray!30}
        }
    }
    \begin{minipage}{0.48\linewidth}
        \centering
        \footnotesize
        \begin{tikzpicture}
            \begin{axis}[
                ymin=0.705, ymax=0.76,
                ylabel={Word Accuracy ($\uparrow$)},
            ]
            
                \addplot [mark=none, blue, thick, domain=5.5:22.5] {0.7422};
                \addlegendentry{Adaptive (Ours)}
                
                \addplot [mark=x, red, thick] coordinates {
                    (6, 0.7213) (8, 0.7247) (10, 0.7358) (12, 0.7236) 
                    (14, 0.7282) (16, 0.7433) (18, 0.7230) (20, 0.7485)
                    (22, 0.7346) 
                };
                \addlegendentry{Fixed Widths}
            \end{axis}
        \end{tikzpicture}
    \end{minipage}
    \hfill
    \begin{minipage}{0.48\linewidth}
        \centering
        \footnotesize
        \begin{tikzpicture}
            \begin{axis}[
                ymin=0.825, ymax=0.87,
                ylabel={NED ($\uparrow$)},
            ]
            
                \addplot [mark=none, blue, thick, domain=5.5:22.5] {0.8598};
                \addlegendentry{Adaptive (Ours)}
                
                \addplot [mark=x, red, thick] coordinates {
                    (6, 0.8432) (8, 0.8473) (10, 0.8501) (12, 0.8461) 
                    (14, 0.8485) (16, 0.8563) (18, 0.8495) (20, 0.8625)
                    (22, 0.8508) 
                };
                \addlegendentry{Fixed Widths}
            \end{axis}
        \end{tikzpicture}
    \end{minipage}
    \caption{\textbf{Ablation of Gaussian widths} in Text-oriented Attention module on the CVTG-Hard (English) subset. We compare the fixed width settings (Red line) against our proposed adaptive Second Central Moment strategy (Blue line). The adaptive method achieves performance competitive with the optimal fixed setting without requiring manual tuning.} 
    \label{fig:sigma_ablation}
\end{figure*}

\noindent \textbf{Ablation on the Gaussian widths.}
To validate the effectiveness of our adaptive strategy, we conduct a comparative analysis on the CVTG-Hard (English) subset.
We compare our proposed adaptive approach against a series of fixed widths:
\textbf{1) Fixed widths.}
We set constant widths $\sigma_{x,k} = \sigma_{y,k}$ for all text instances regardless of their size, testing values in $\{6, 8, \dots, 22\}$.
\textbf{2) Second central moment (Adaptive).}
We adopt the adaptive strategy described in Section~\ref{sec:attn}, where the Gaussian widths are dynamically computed as the second central moments of the smoothed attention map, allowing the gate to adapt to the text's actual coverage without manual tuning.
The quantitative results are reported in Figure~\ref{fig:sigma_ablation}. 
Fixed widths vary substantially across settings, indicating strong sensitivity to parameter choice. 
In contrast, the proposed second central moment consistently approaches the best-performing fixed setting. 
This attention-driven formulation is adaptive and eliminates the need for manual hyperparameter tuning. We therefore use it in all subsequent experiments.


\noindent \textbf{Complexity Analysis.}
\yt{In TextCrafter~(Qwen-Image) at $1024 \times 1024$ resolution, the Text Insulation module incurs no additional inference latency.
The introduction of Text-oriented Attention increases the peak GPU memory usage from $\sim$55GB to $\sim$60GB and extends inference time due to extra attention computations.
However, our framework is compatible with distillation techniques such as DMD2~\cite{yin2024improved} and sCM~\cite{lusimplifying}. This integration can reduce the number of sampling steps from $50$ to $8$ (or even $4$), delivering over a $6\times$ speedup while enabling controllable quality–efficiency trade-offs.
}

%

%


\section{Conclusion}
\yt{
In this paper, we present TextCrafter, a Complex Visual Text Generation (CVTG) framework inspired by selective visual attention in cognitive science.
TextCrafter employs the ``Text Insulation-and-Attention" mechanisms to address the challenges of text misgeneration, omissions, and hallucinations. 
Specifically, on one hand, TextCrafter presents a Text Insulation module that introduces a novel Bottleneck-aware Constrained Reinforcement Learning method to explicitly optimize the fidelity of each text instance.
On the other hand, TextCrafter introduces a text-oriented attention module with a novel Quotation-guided Attention Gate to further enhance text rendering.
Last but not least, we introduce CVTG-2K, the first challenging benchmark for complex visual text generation, comprising 2,000 complex visual-text prompts.
Extensive experiments demonstrate that TextCrafter significantly outperforms existing state-of-the-art models on several challenging datasets: CVTG-2K, CVTG-Hard and LongText-Bench, including highly capable industry models (\emph{e.g.}, GPT Image, Qwen-Image, Seedream \emph{etc.}) that are trained with substantial resources.
}

\vspace{1mm}
\noindent \textbf{Limitations.}
\yt{
An effective text rendering model should not only generate text accurately but also minimize \emph{hallucinated content}. 
Although we incorporate the ``Insulation and Attention" mechanisms to improve text rendering performance, 
it remains challenging for current state-of-the-art industrial systems and our TextCrafter to ensure the complete absence of hallucinations or textual inaccuracies within a single generation. Consequently, enhancing the robustness of complex text rendering warrants further in-depth exploration.
}

\section*{Acknowledgment}
This work was supported by the Gusu Innovation and Entrepreneur Leading Talents: No. ZXL2024362, Natural Science Foundation of Jiangsu Province: BK20241198, and Natural Science Foundation of China: No. 62406135.

\section*{Data Availability Statement}
CVTG-2K dataset is available at: \url{https://huggingface.co/datasets/dnkdnk/CVTG-2K}

\noindent MixGRPO training dataset is available at: 
\url{https://github.com/Tencent-Hunyuan/MixGRPO/blob/main/data/prompts.txt}

\noindent Qwen-Image-Self-Generated-Dataset is available at: \url{https://modelscope.cn/datasets/DiffSynth-Studio/Qwen-Image-Self-Generated-Dataset}

\bibliography{references} 

\clearpage
\appendix
\section*{Supplementary Material}
\section{Prompt Design of O1-mini}\label{promptdesign}
\begin{figure*}[t!]
  \centering
  \includegraphics[width=0.7\textwidth]{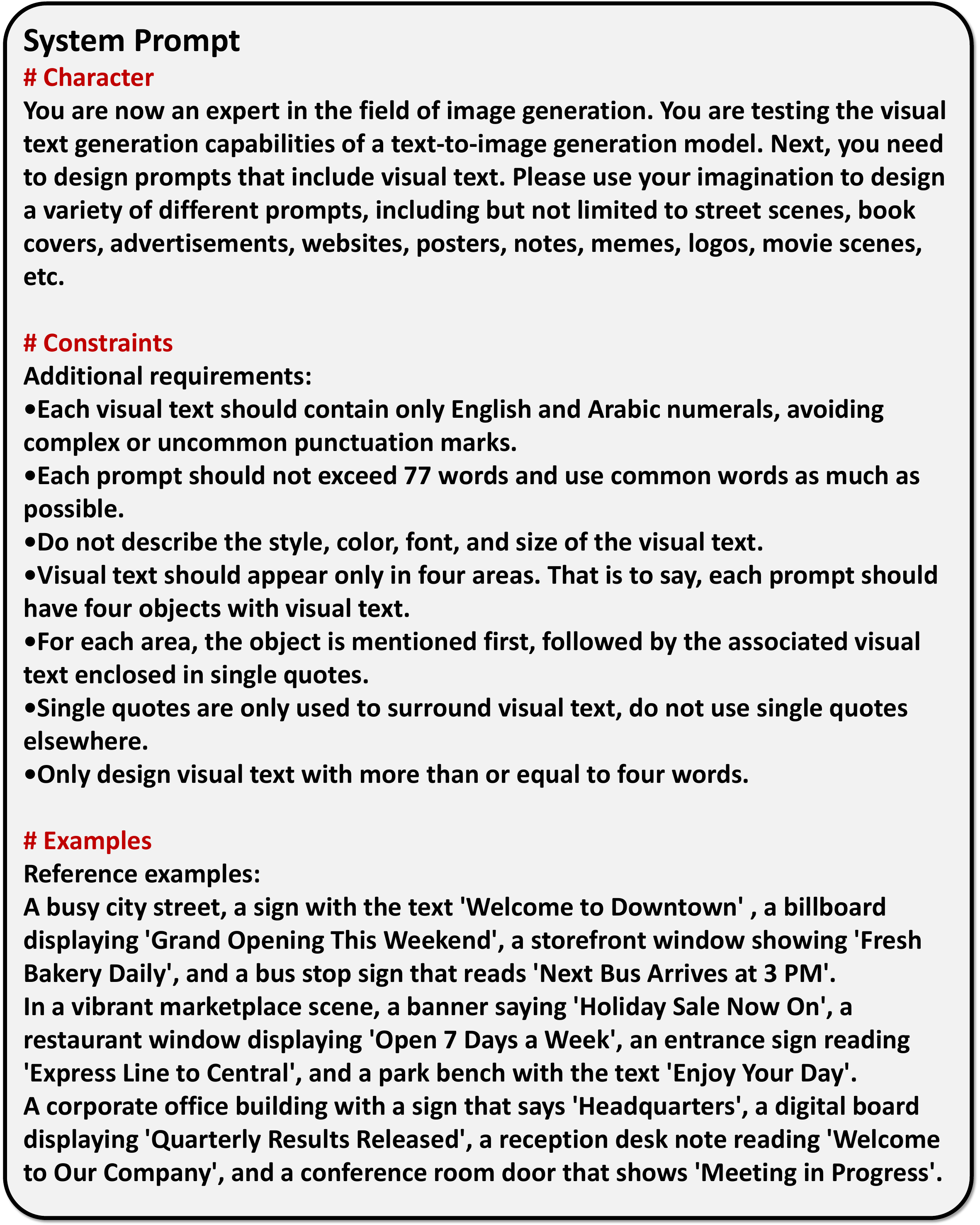}
  \caption{System prompt of original CVTG-2K.}
  \label{fig:prompt}
\end{figure*}

We divided the system prompt into three parts (\emph{i.e.}, Character, Constraints and Examples), aiming to both impose constraints on the model and stimulate its creativity. Through extensive testing on challenging samples, including numerous visual texts, uncommon words, and complex scenes, we finalized the system prompt shown in Figure~\ref{fig:prompt}. 
We designed prompts to generate varying numbers of regions and word counts, with the number of regions ranging from 2 to 5, and word counts categorized into single words (short), 2 or 3 words (medium length), and more than 4 words (long). Figure~\ref{fig:prompt} illustrates an example with 3 regions and medium-length visual texts.

\begin{figure*}[t!]
    \centering
    \includegraphics[width=0.75\textwidth]{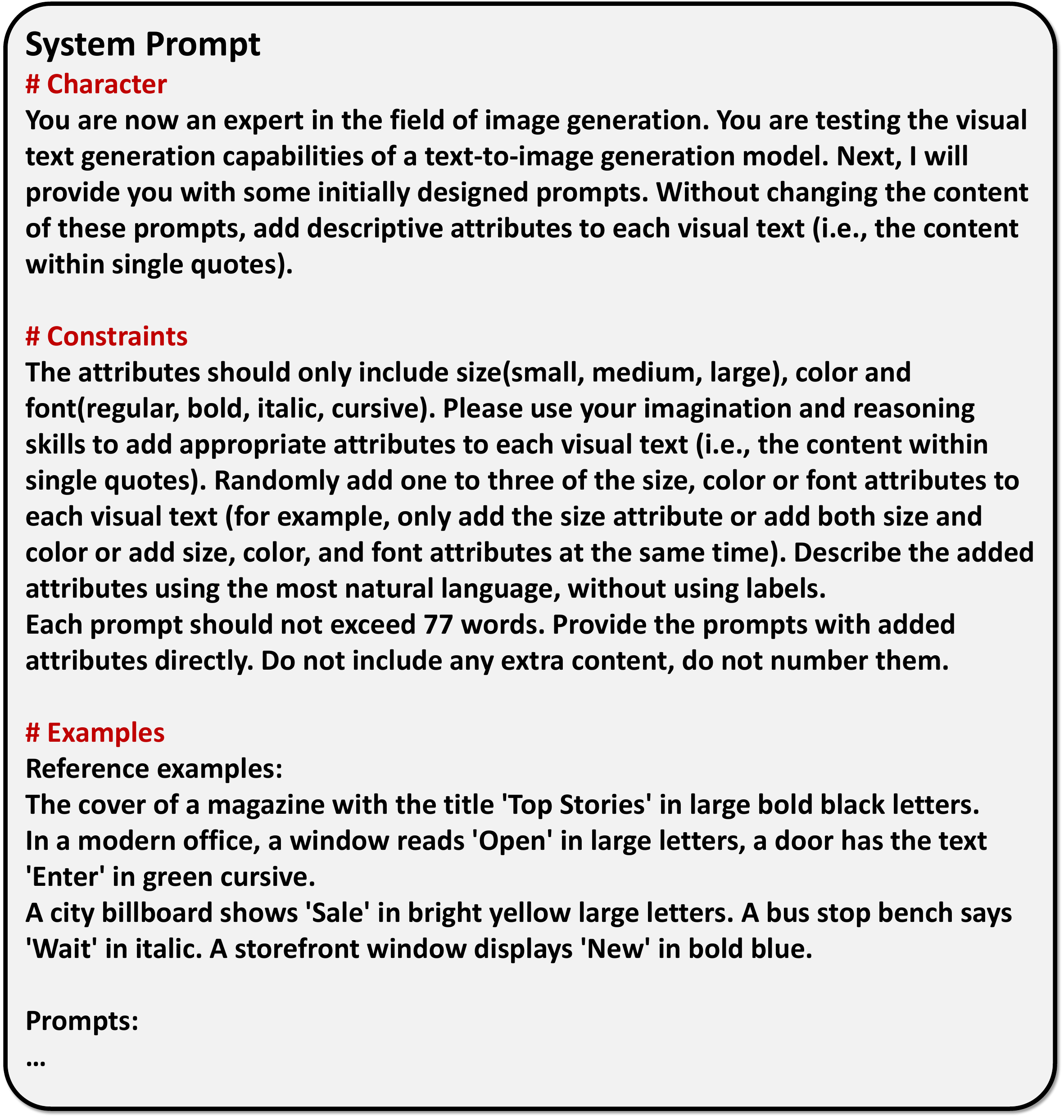}
    \caption{System prompt of inserting attributes to CVTG-2K.}
    \label{attribute}
\end{figure*}

\noindent \textbf{Attribute Insertion.} After generating the original benchmark, we selected \emph{half} of the samples to introduce additional attributes, aiming to test the image generation model's ability to control visual text attributes. We also divided the system prompt into three parts to maximize the language model's reasoning capabilities. The attributes to be added are limited to three types: size, color, and font, with each visual text randomly assigned one or more attributes. Figure~\ref{attribute} illustrates the system prompt used for adding attributes.

\begin{figure*}[t!]
    \centering
    \includegraphics[width=0.75\textwidth]{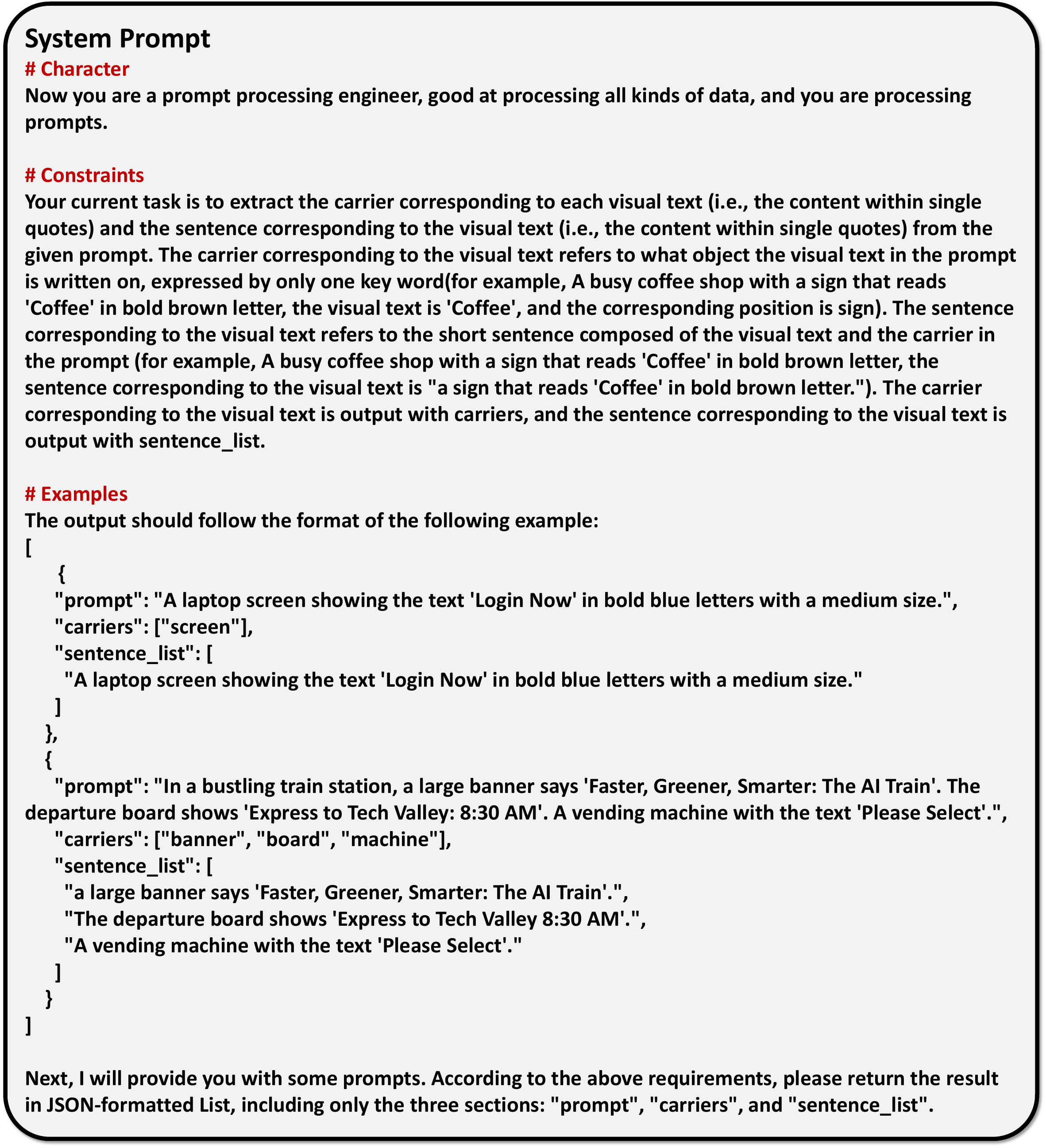}
    \caption{System prompt of fine-grained information.}
    \label{decouple}
\end{figure*}

\noindent \textbf{Fine-grained Information.} Additionally, we annotated more fine-grained information for CVTG-2K to assist future research on the CVTG task. We decoupled complex multiple visual texts, isolated clauses that describe each visual text, and used a key word to represent the carrier of the visual text. The results are stored in JSON format. Figure~\ref{decouple} illustrates the system prompt used for annotating granular information.

\section{Example of CVTG-2K}\label{example}
\begin{figure*}[t!]
    \centering
    \includegraphics[width=\textwidth]{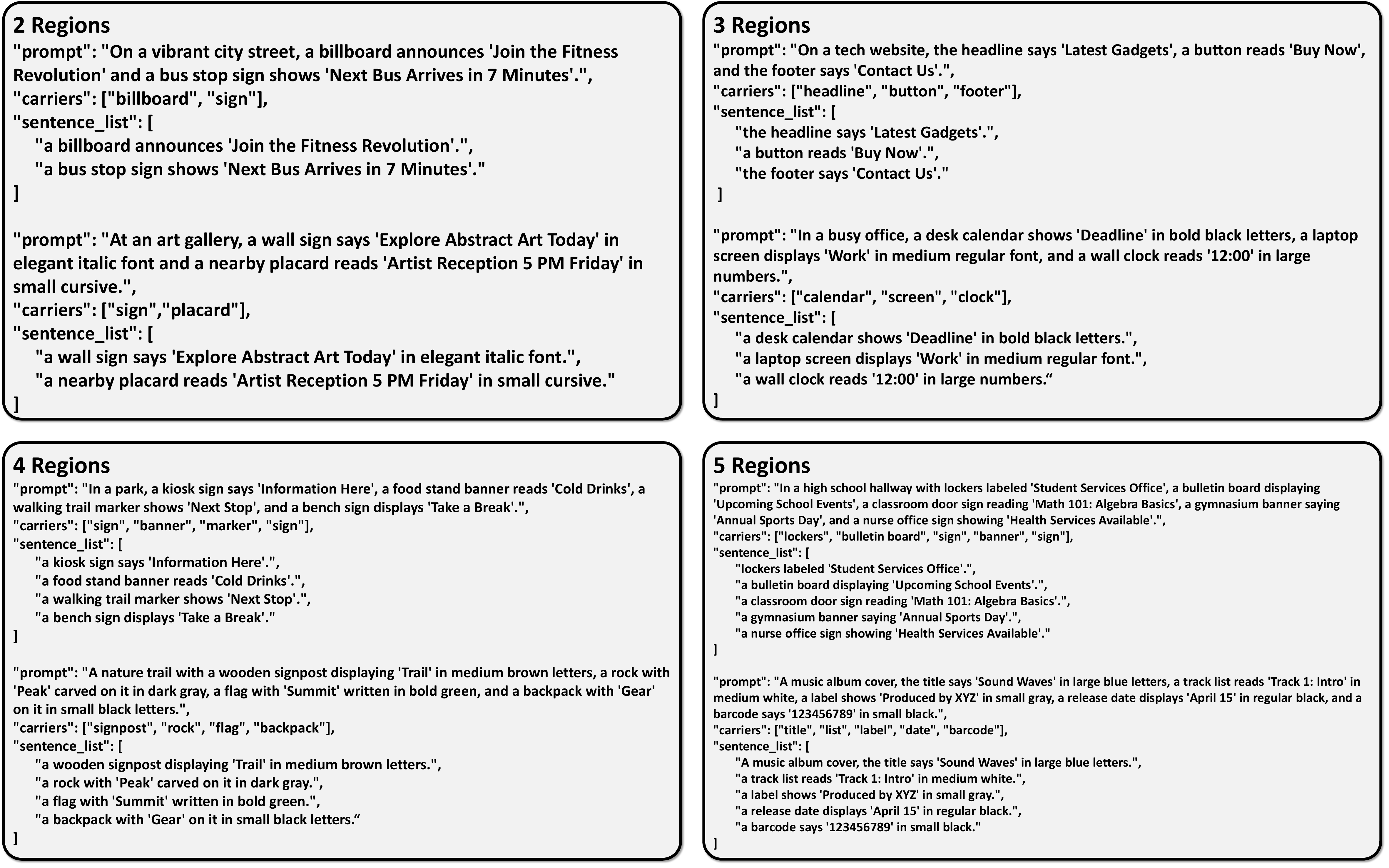}
    \caption{Randomly selected samples under different numbers of regions.}
    \label{samples}
\end{figure*}

Using the method introduced in \textbf{Prompt Design of O1-mini}, we assign different proportions to different numbers of regions. From 2 regions to 5 regions, the proportions are approximately 20\%, 30\%, 30\%, and 20\% respectively. We also assign different proportions to texts of different lengths, with medium-length texts accounting for the largest number, accounting for approximately 60\%. Figure~\ref{samples} shows randomly selected samples under different numbers of regions.

\section{Detailed Configurations of the Metric Calculation}\label{quantitative}

\noindent\textbf{Notation and Preprocessing.}
For each prompt $P_j$, we use the same symbol definition as the main paper and denote the target visual-text set as
$VT_j=\{vt_{j,1},vt_{j,2},\dots,vt_{j,n_j}\}$.
The evaluator extracts $VT_j$ by collecting all strings enclosed by single quotes in $P_j$, then lowercasing and splitting by whitespace.
Given image $I_j$, PPOCR-v4 outputs an OCR word list $\mathcal{O}_j$.
For each ground-truth text $vt_{j,i}$, the matched OCR word is selected by top-1 \texttt{difflib.get\_close\_matches}:
\begin{equation}
\hat{vt}_{j,i}=\operatorname{Match}(vt_{j,i},\mathcal{O}_j).
\end{equation}

\noindent\textbf{Word Accuracy.}
At region count $r$ (sample set $\mathcal{S}_r$), a target word is counted as correct only if an exact match appears in $\mathcal{O}_j$:
\begin{equation}
\operatorname{WordAcc}_{r}=\frac{\sum_{j\in\mathcal{S}_r}\sum_{i=1}^{n_j}\mathbf{1}\left[vt_{j,i}\in\mathcal{O}_j\right]}{\sum_{j\in\mathcal{S}_r}n_j}.
\end{equation}

\noindent\textbf{Normalized Edit Distance (NED).}
Consistent with the script, NED is implemented as the normalized edit \emph{similarity} ($\uparrow$). For a subset $\mathcal{S}_r$:
\begin{equation}
\resizebox{\linewidth}{!}{%
$
\operatorname{NED}_{r}=\frac{1}{\sum_{j\in\mathcal{S}_r}n_j}\sum_{j\in\mathcal{S}_r}\sum_{i=1}^{n_j}
\left(1-\frac{\operatorname{dist}(vt_{j,i},\hat{vt}_{j,i})}{\max\left(|vt_{j,i}|,|\hat{vt}_{j,i}|\right)+\epsilon}\right),\; \epsilon=10^{-5}
$%
}
\end{equation}
where $\operatorname{dist}(\cdot,\cdot)$ is the Levenshtein distance.

\noindent\textbf{CLIPScore.}
The input text is prefixed as $T_j=\text{``A photo depicts ''}+P_j$.
Using ViT-L/14 features with $L_2$ normalization ($f(\cdot)$), the score for image $I_j$ is:
\begin{equation}
\operatorname{CLIPScore}(I_j, P_j)=2.5\cdot\max\left(0, f(I_j)^\top f(T_j)\right).
\end{equation}
The region-level result is the average over images:
\begin{equation}
\operatorname{CLIPScore}_{r}=\frac{1}{|\mathcal{S}_r|}\sum_{j\in\mathcal{S}_r}\operatorname{CLIPScore}(I_j, P_j).
\end{equation}

\noindent\textbf{VQAScore and Aesthetics.}
VQAScore is computed by \texttt{t2v\_metrics} (\texttt{clip-flant5-xxl}) and Aesthetics by LAION Aesthetics Predictor V1:
\begin{equation}
\operatorname{VQAScore}_{r}=\frac{1}{|\mathcal{S}_r|}\sum_{j\in\mathcal{S}_r}\operatorname{VQA}(I_j,P_j),
\end{equation}
\begin{equation}
\operatorname{Aes}_{r}=\frac{1}{|\mathcal{S}_r|}\sum_{j\in\mathcal{S}_r}\operatorname{Aes}(I_j).
\end{equation}

\noindent\textbf{Overall Aggregation.}
For the final summary over all benchmark subsets/region groups $u$, Word Accuracy and NED are micro-averaged by total word count:
\begin{equation}
\begin{split}
\operatorname{WordAcc}_{\text{all}} &= \frac{\sum_u C_u}{\sum_u N_u}, \\
\operatorname{NED}_{\text{all}} &= \frac{\sum_u\sum_{k=1}^{N_u}\operatorname{ned}_{u,k}}{\sum_u N_u}.
\end{split}
\end{equation}
where $C_u$ is the count of correct words and $N_u$ is the total number of words in subset $u$.
Conversely, CLIPScore, VQAScore, and Aesthetics are averaged by image count:
\begin{equation}
\begin{split}
M_{\text{all}} &= \frac{\sum_u |\mathcal{S}_u|\cdot M_u}{\sum_u |\mathcal{S}_u|}, \\
M &\in\{\operatorname{CLIPScore},\operatorname{VQAScore},\operatorname{Aes}\}.
\end{split}
\end{equation}

\section{Visual Comparison with Industrial Models}\label{sec:industrial_comparison}
\begin{figure*}[t]
    \centering
    \includegraphics[width=0.75\textwidth]{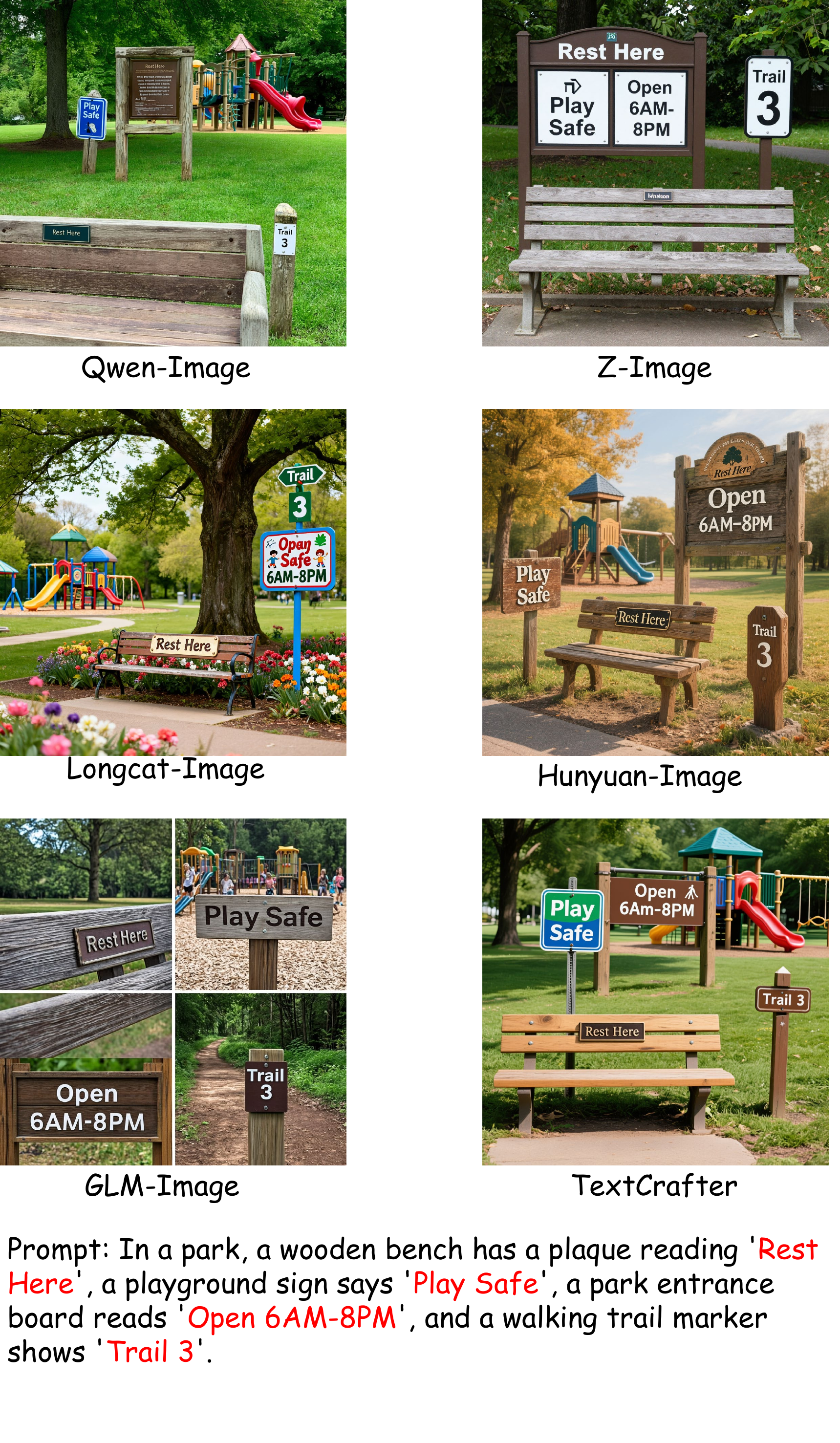}
    \caption{Visual comparison on CVTG-Hard (Sample 1). In the prompt, \textcolor{red}{red} indicates the target visual text, and \textcolor{blue}{blue} indicates the required attributes.}
    \label{fig:supp_qual_1}
\end{figure*}

\begin{figure*}[t]
    \centering
    \includegraphics[width=0.75\textwidth]{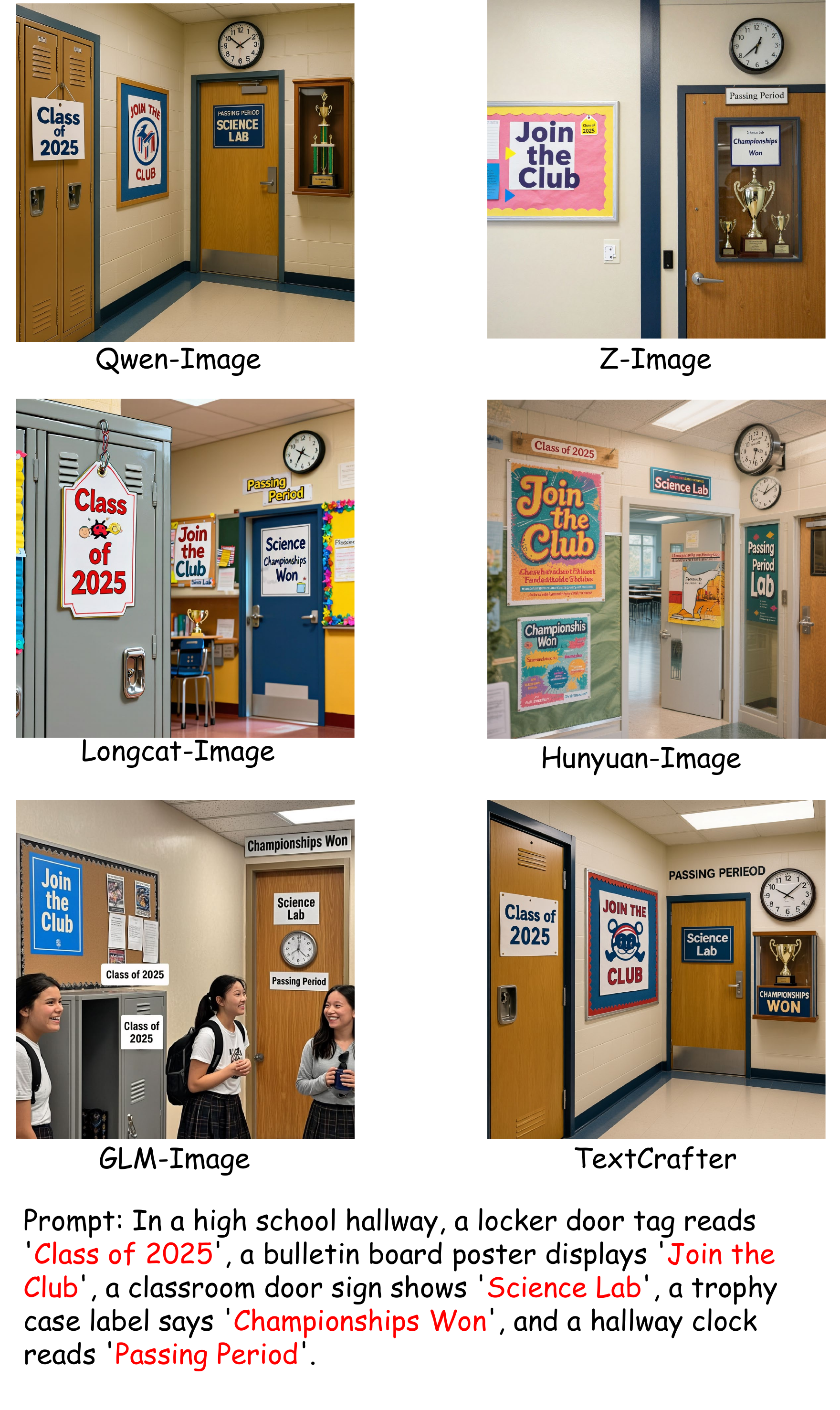}
    \caption{Visual comparison on CVTG-Hard (Sample 2). In the prompt, \textcolor{red}{red} indicates the target visual text, and \textcolor{blue}{blue} indicates the required attributes.}
    \label{fig:supp_qual_2}
\end{figure*}

\begin{figure*}[t]
    \centering
    \includegraphics[width=0.75\textwidth]{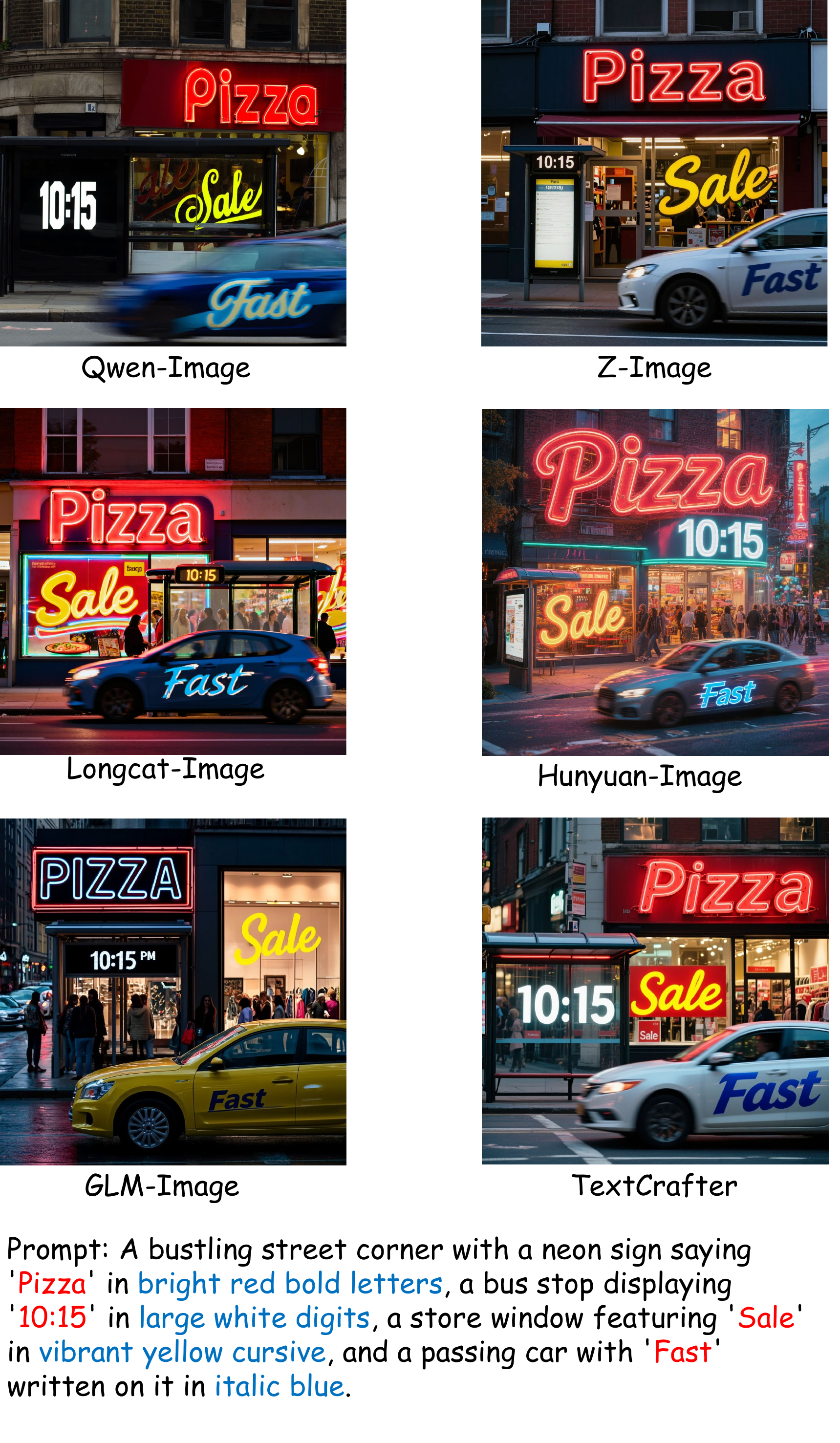}
    \caption{Visual comparison on CVTG-Hard (Sample 3). In the prompt, \textcolor{red}{red} indicates the target visual text, and \textcolor{blue}{blue} indicates the required attributes.}
    \label{fig:supp_qual_4}
\end{figure*}

\begin{figure*}[t]
    \centering
    \includegraphics[width=0.75\textwidth]{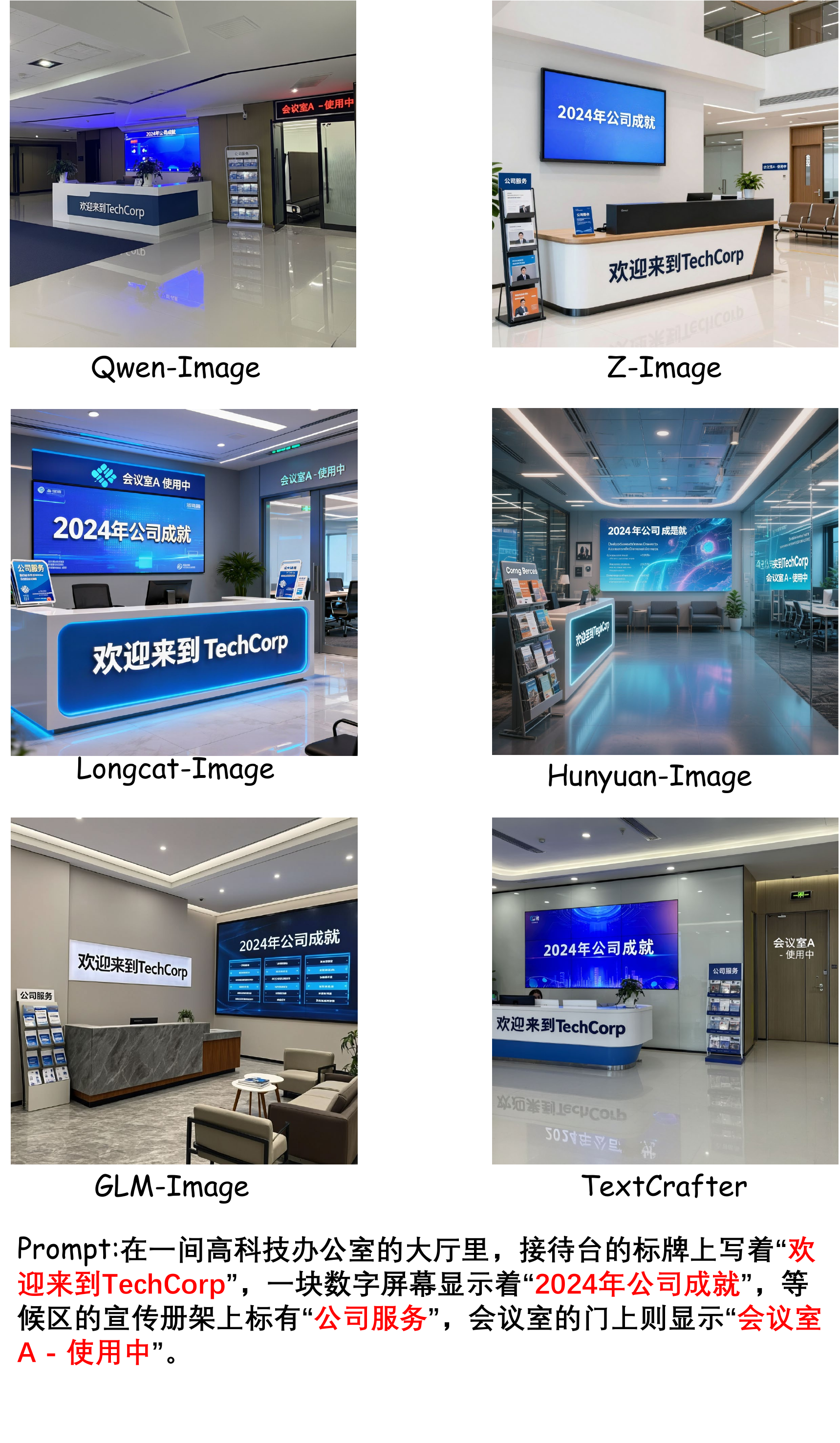}
    \caption{Visual comparison on CVTG-Hard (Sample 4). In the prompt, \textcolor{red}{red} indicates the target visual text, and \textcolor{blue}{blue} indicates the required attributes.}
    \label{fig:supp_qual_6}
\end{figure*}

\begin{figure*}[t]
    \centering
    \includegraphics[width=0.75\textwidth]{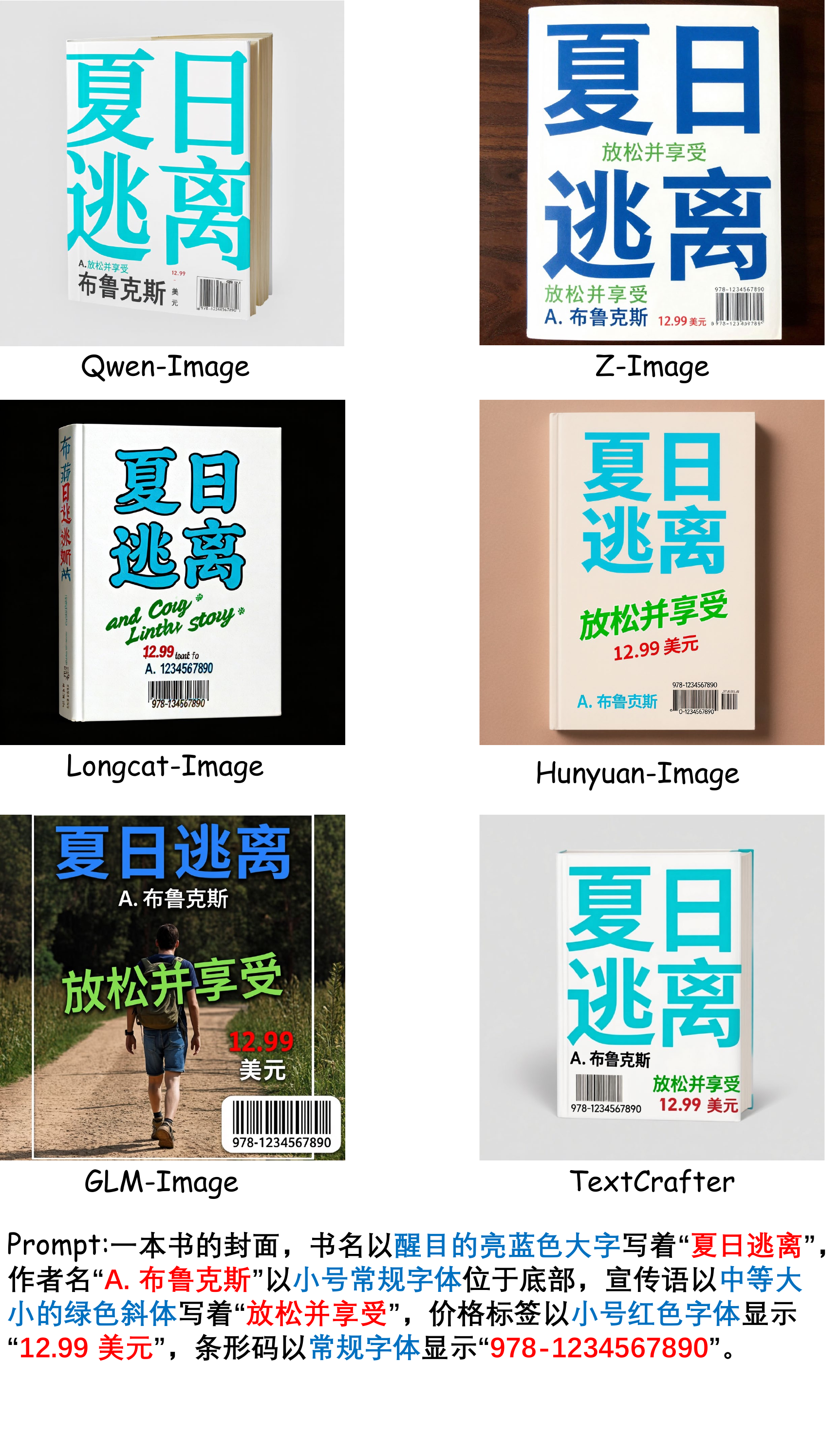}
    \caption{Visual comparison on CVTG-Hard (Sample 5). In the prompt, \textcolor{red}{red} indicates the target visual text, and \textcolor{blue}{blue} indicates the required attributes.}
    \label{fig:supp_qual_7}
\end{figure*}

\begin{figure*}[t]
    \centering
    \includegraphics[width=0.75\textwidth]{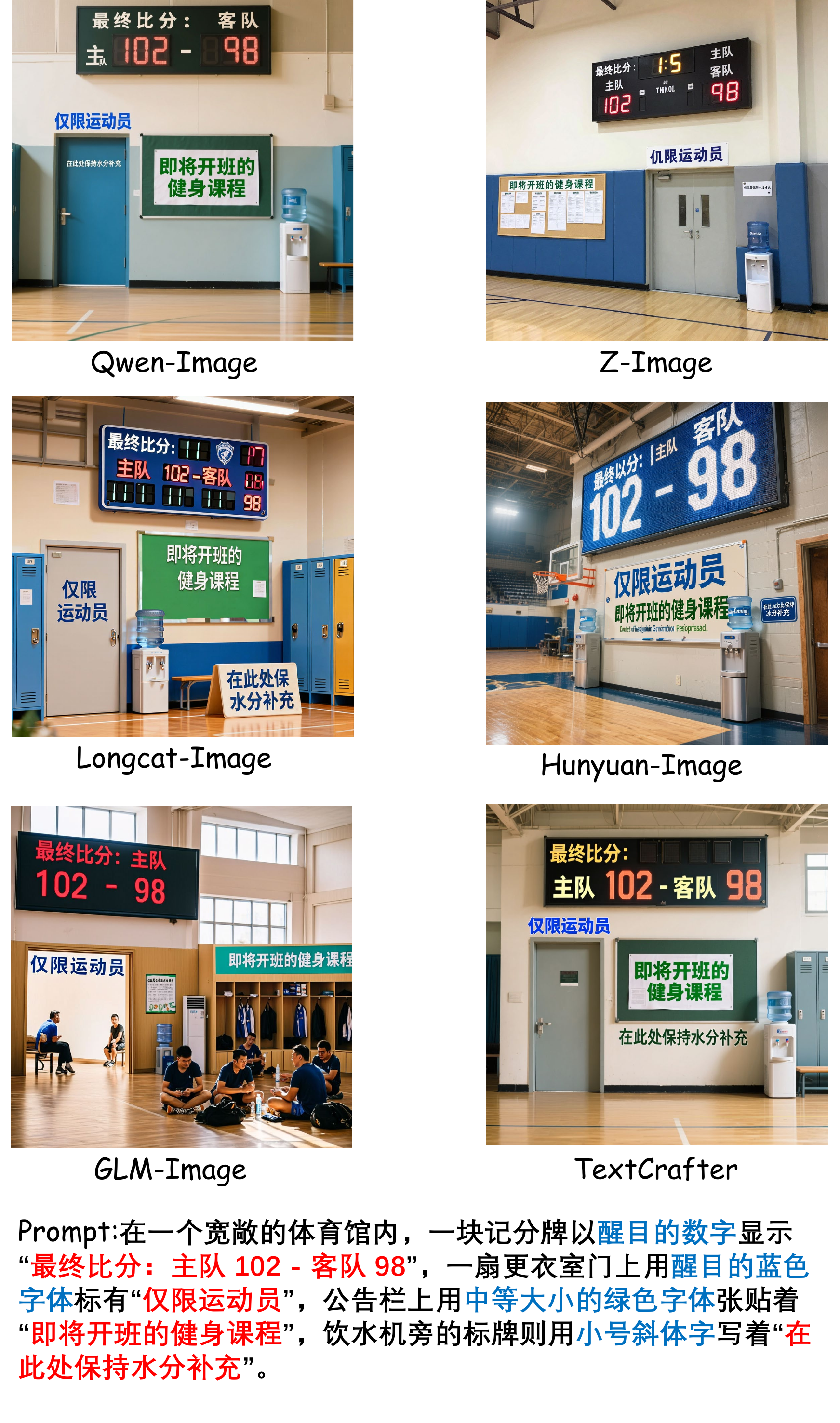}
    \caption{Visual comparison on CVTG-Hard (Sample 6). In the prompt, \textcolor{red}{red} indicates the target visual text, and \textcolor{blue}{blue} indicates the required attributes.}
    \label{fig:supp_qual_8}
\end{figure*}

In this section, we present a comprehensive visual comparison between TextCrafter and state-of-the-art industrial text-to-image models, including Qwen-Image, Z-Image, Longcat-Image, Hunyuan-Image, and GLM-Image. 
All comparisons are conducted on the challenging \textbf{CVTG-Hard} subset proposed in our main paper, which involves complex layouts and rich attribute constraints. 
Figures~\ref{fig:supp_qual_1}-\ref{fig:supp_qual_8} illustrate the qualitative results.
For clarity in the prompts shown in the figures, the \textcolor{red}{red content} indicates the required visual text, while the \textcolor{blue}{blue content} denotes the specific attributes (\emph{e.g.}, color, font, style) required for the text.
As demonstrated, TextCrafter consistently outperforms these powerful industrial models, particularly in text accuracy and attribute adherence.

\end{document}